\documentclass[letterpaper]{article} 
\usepackage{aaai2026}  
\usepackage{times}  
\usepackage{helvet}  
\usepackage{courier}  
\usepackage[hyphens]{url}  
\usepackage{graphicx} 
\usepackage{natbib}  
\usepackage{caption} 
\usepackage{algorithm}
\usepackage{algorithmic}
\usepackage{booktabs}        
\usepackage{multirow}        
\usepackage{siunitx}         
\usepackage{caption}
\usepackage{amsmath} 
\usepackage{graphicx}
\usepackage{subcaption}
\usepackage{amssymb}
\usepackage{fontawesome}
\usepackage{url} 
\usepackage{xcolor}

\makeatletter
\def\@fnsymbol#1{%
  \ifcase#1\or
    *\or                           
    $\dagger$\or                   
    \faEnvelopeO\else              
    \@ctrerr                       
  \fi
}
\makeatother
\usepackage{newfloat}
\usepackage{listings}
\DeclareCaptionStyle{ruled}{labelfont=normalfont,labelsep=colon,strut=off} 
\floatstyle{ruled}
\newfloat{listing}{tb}{lst}{}
\floatname{listing}{Listing}
\title{CURE: Critical-Token-Guided Re-Concatenation \\for
Entropy-Collapse Prevention}

\author{
    Qingbin Li$^{\heartsuit}$\thanks{Equal Contribution.}\thanks{Work done during internship at ByteDance.} 
    Rongkun Xue$^{\blacklozenge}$\footnotemark[1]\footnotemark[2], 
    Jie Wang$^{\heartsuit}$, 
    Ming Zhou$^{\clubsuit}$, 
    Zhi Li$^{\heartsuit}$
    Xiaofeng Ji$^{\heartsuit}$,\\ 
    Yongqi Wang$^{\heartsuit}$, 
    Miao Liu$^{\heartsuit}$, 
    Zheming Yang$^{\heartsuit}$, 
    Minghui Qiu$^{\heartsuit}$\thanks{Corresponding authors.}, 
    Jing Yang$^{\blacklozenge}$\footnotemark[3]
}

\affiliations{
      $^{\heartsuit}$ByteDance 
    $^{\blacklozenge}$ Xi’an Jiaotong University
    $^{\clubsuit}$Shanghai AI Laboratoy\\

}

\usepackage{bibentry}

\begin{document}
\nocopyright
\maketitle

\begin{abstract}
Recent advances in Reinforcement Learning with Verified Reward (RLVR) have driven the emergence of more sophisticated cognitive behaviors in large language models (LLMs), thereby enhancing their reasoning capabilities. However, in prior RLVR pipelines, the repeated use of static initial-state sampling drawn exactly from the dataset distribution during each sampling phase produced overly deterministic, low diversity model behavior, which manifested as rapid entropy collapse and hindered sustained performance gains during prolonged training. To address this issue, we introduce \textbf{CURE} (\textbf{C}ritical-token-g\textbf{U}ided \textbf{R}e concatenation for \textbf{E}ntropy-collapse prevention), a two-stage framework that balances exploration and exploitation. Specifically, in the first stage, to deliberately steer the model toward novel yet coherent contexts, we re-generate at high-entropy critical tokens and jointly optimize the original and the branched trajectories. The further comparison with vanilla DAPO shows that the regeneration process achieves a better performance on math reasoning tasks while sustaining a high-level entropy degree for exploration. In the second stage, we continue training with static initial-state sampling by DAPO, intentionally placing the model in a familiar state to gradually strengthen exploitation. Extensive experiments on Qwen-2.5-Math-7B show that, compared to other RLVR methods, CURE achieves a $5$\% performance gain across six math benchmarks, establishing state-of-the-art performance in both entropy and accuracy. A series of experiments further validate the effectiveness of our approach. Code is available at https://github.com/bytedance/CURE.

\end{abstract}
\section{Introduction}
Recent advancements in Reinforcement Learning with Verification (RLVR) ~\cite{jaech2024openai,zeng2025simplerl,hu2025open,liu2025understanding} have driven significant progress in unlocking the reasoning capabilities of large language models. By replacing opaque reward surrogates with automatic verifiers that emit precise binary signals~\cite{deepseekai2025deepseekr1incentivizingreasoningcapability}, RLVR enables scalable, self-improving training loops and has delivered strong gains on challenging reasoning benchmarks, from mathematical problem solving~\cite{dataset_math,dataset_olympiad} to scientific QA~\cite{rein2024gpqa}.

\begin{figure}[t]
\small
\vspace{3mm}
\centering
\includegraphics[width = \linewidth]{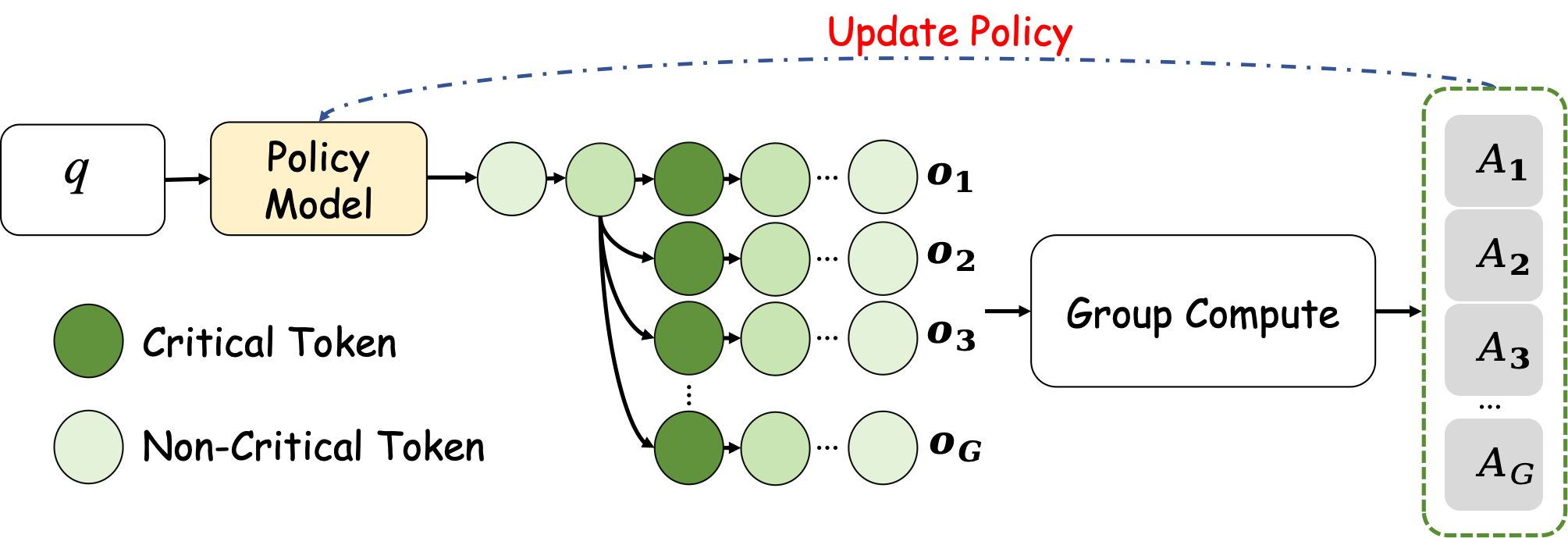}
\vspace{-3mm}
\caption{Overview of the CURE pipeline. In Stage 1, given an input query 
$q$, the policy model produces a pool of candidate responses. We compute token-level entropy to identify critical tokens (high entropy), extract the clauses immediately preceding those tokens, append them to 
$q$ to form refined prompts, and query the model again. The newly generated responses are aggregated with the original ones and jointly optimized within a single group. In Stage 2, we continue training to translate the exploration bonus into realized performance.}
\label{fig:main}
\end{figure}

Despite the impressive gains of RLVR, the community~\cite{cui2025entropy,tarvainen2017mean} now recognizes policy entropy collapse as the key bottleneck blocking further progress. Once entropy collapses, probability mass concentrates on a few low-diversity responses, and performance plateaus early. Early—often ad-hoc—fixes include simply raising the sampling temperature~\cite{zhang2025srpo} and adding a small KL term~\cite{deepseekai2025deepseekr1incentivizingreasoningcapability,zhou2025r1} to slow the drop. 
More principled attempts add entropy regularizers, redesign the loss, or reward shaping~\cite{liu2025prorl,cheng2025reasoning}.
Furthermore, ProRL~\cite{liu2025prorl} applies reference‐policy resets on a 1.5B‐parameter model by periodically hard‐resetting the reference policy to a recent snapshot of the online policy to curb entropy decay. However, it demands frequent reference‐model updates and optimizer resets.
Recently, Clip-Cov~\cite{cui2025entropy} shows that clipping high-covariance tokens can also help prevent entropy collapse and prolong stable training, though it still requires fine-grained, task-specific hyperparameter tuning, yields only limited performance improvements.
Ultimately, the need for heavy hyperparameter tuning and manual model updates arises because prior work keeps adjusting the update rule while still relying on a fixed set of prompts and verifier signals. A static training-state distribution quickly narrows the exploration space, so entropy collapse is almost guaranteed.

\begin{figure*}[t]
\small
\vspace{3mm}
\centering
\includegraphics[width = 1.0\linewidth]{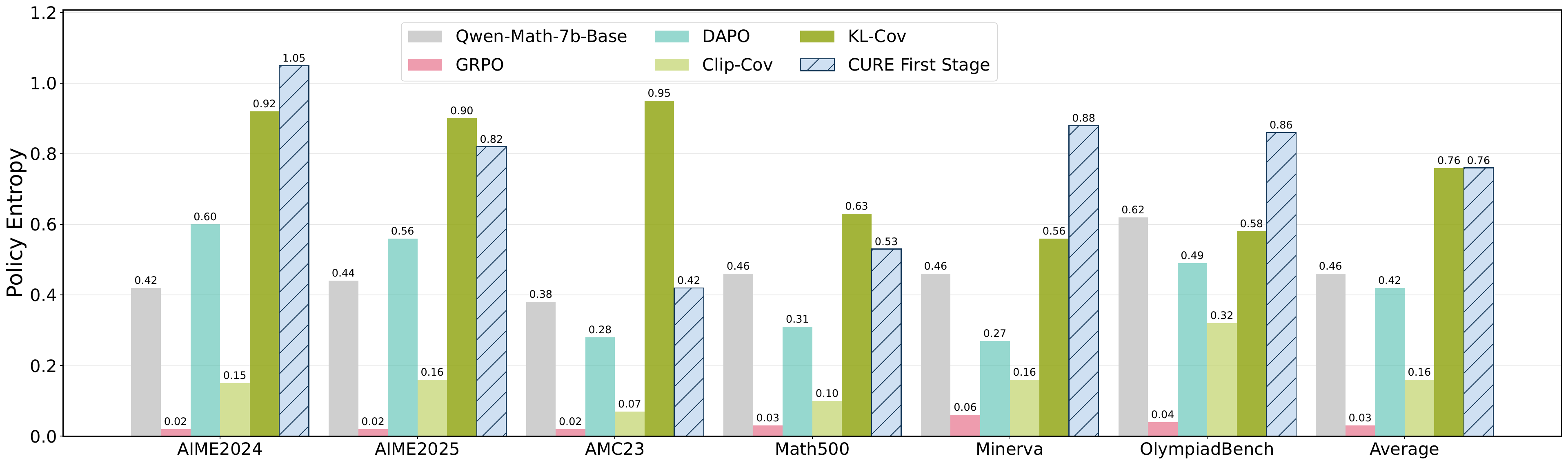}
\caption{Entropy comparison of CURE first stage and other methods at temperature 1.0}
\label{fig:model_relase_1}
\end{figure*}

We argue that an effective and conceptually simple strategy is to continually exploit uncertainty during response generation, thereby preventing entropy collapse~\cite{ladosz2022exploration,ecoffet2019go,burda2018exploration}. We dynamically enrich the training signal by leveraging the model’s own internal uncertainty---specifically, token-level policy entropy computed during autoregressive decoding. High-entropy tokens expose moments of genuine indecision. A naive tactic is to append a randomly sampled continuation to the original query so that subsequent rollouts start from a richer prompt. In practice, however, this inflates the context with irrelevant detail and still fails to probe the policy’s true blind spots. Our remedy is to intervene exactly at the decision point where the model is most uncertain. For each generated answer, we locate the token with the highest policy entropy and truncate the sequence immediately before it. The retained prefix marks a high-stakes fork in the reasoning process. From this fork, the current policy produces multiple alternative continuations. 
Because these updates act precisely where the model was undecided, they simultaneously broaden the distribution of plausible next moves---maintaining healthy entropy and raise expected accuracy in future rollouts.

Building on these insights, we translate uncertainty‑aware exploration into a concrete training routine that broadens the state distribution at critical decision points before consolidating gains. In this work, we propose \textbf{CURE} (\textbf{C}ritical-token-g\textbf{U}ided \textbf{R}e-concatenation for \textbf{E}ntropy-collapse prevention), a two-stage RLVR framework designed to balance exploration and exploitation by dynamically expanding the state distribution during training.
Specifically, in the first stage, we first identify critical tokens in each response using a token‑level entropy criterion; next, we re‑concatenate only the clauses that precede these critical tokens with the original query to create a refined prompt; finally, we feed  prompt back into the model, thereby encouraging exploration along trajectories that are conditioned on previously unseen yet semantically salient context. To improve training efficiency, we follow DAPO~\cite{yu2025dapo} by retaining only those prompts whose candidate responses include both verifier‑accepted (correct) and verifier‑rejected (incorrect) answers. In the second stage, CURE transitions from high-entropy exploration to low-entropy exploitation through a continue training process. Specifically, we revert to static initial-state sampling using the DAPO~\cite{yu2025dapo} strategy, deliberately placing the model back into familiar prompt contexts. This shift encourages the policy to consolidate knowledge gained during the exploratory phase and reinforce accurate behaviors in a more deterministic regime. Crucially, while exploration-focused re-concatenation is disabled in this stage, the model still benefits from the enriched policy distribution developed earlier. As training proceeds under a static query distribution, the reward naturally increases, which helps sharpen decision boundaries and stabilize convergence. This two-phase training regime ensures that the policy first explores broadly and then exploits precisely, maintaining strong generalization while achieving good performance on verifier-based reasoning tasks.


Our contributions can be summarized as follows:

\begin{itemize}
\item We present the first analysis of policy entropy collapse from the perspective of state distribution, showing that training on a fixed dataset depresses entropy; we further provide a principled remedy based on critical-token-guided re-concatenation.
\item We propose CURE, a lightweight two-stage framework that balances exploration and exploitation by shaping the training-state distribution. 

\item We perform extensive experiments across several challenging math‑reasoning benchmarks. CURE not only achieves state‑of‑the‑art accuracy but also retains the highest entropy among similarly performing models, underscoring its capacity for continued improvements.
\end{itemize}

\section{Related Work}
\paragraph{Reinforcement Learning for LLM Reasoning.}
Early approaches to reinforcement learning for large language models leveraged reward-model-based fine-tuning and direct preference optimization~\cite{rafailov2023direct,christiano2017deep,ouyang2022training,Cui2023ULTRAFEEDBACKBL}, simulating human feedback to guide model behavior. However, these methods depended heavily on manually collected human preferences or synthetic preference models~\cite{Yuan2024AdvancingLR,wei2024magicoder,liu2024acemath}.
More recently, the community has shifted toward pure RL at scale for reasoning. Openai o1~\cite{jaech2024openai} first showed that large-scale RL trained solely with scalar outcome rewards can yield significant reasoning gains in LLMs. Subsequent studies~\cite{deepseekai2025deepseekr1incentivizingreasoningcapability,team2025kimi,cui2025process,zeng2025simplerl,hu2025open,liu2025understanding,yan2025learning,chen2025bridging} empirically validated that a minimal RL recipe—using only outcome signals and no auxiliary preference models—scales effectively and unlocks further performance improvements.
\vspace{-5pt}
\paragraph{Entropy Control.}
Policy entropy collapse constitutes a critical obstacle in reinforcement learning (RL) for large language models (LLMs), as diminishing exploration often leads to rapid performance stagnation.
Regularization-based approaches~\cite{he2025skywork,liu2025prorl} add an entropy loss or KL loss whose weight often needs careful or adaptive tuning. Reward shaping methods~\cite{cheng2025reasoning} similarly adds an entropy bonus to the reward or advantage to balance exploration and exploitation. Complementary interventions like loss reweighting~\cite{wang2025beyond,cui2025entropy} and clip-higher~\cite{yu2025dapo} can also help to prevent entropy collapse.

In contrast to objective-shaping and static-sampling regularizers, CURE is a two-stage data-level framework: it leverages high-entropy critical tokens to drive exploration, then continues training with DAPO for exploitation, improving the exploration-exploitation trade-off.

\section{Method}
\subsection{Background}
\subsubsection{MDP Formulation of Language Generation.}
We formalize token-level language generation as a Markov Decision Process (MDP) \(\mathcal{M}=(\mathcal{S},\mathcal{A},\mathbb{P},R,d_{0},\omega)\) with fully observable, deterministic dynamics. For a prompt \(\mathbf{q}\sim d_{0}\) with \(\mathbf{q}=(q_{0},\dots,q_{m})\), we sample \(G\) rollouts from the behavior policy \(\pi_{\theta_{\mathrm{old}}}\). Rollout \(i\in\{1,\dots,G\}\) is the token sequence \(\mathbf{o}_{i}=(o_{i,1},\dots,o_{i,T_i})\in\mathcal{V}^{T_i}\) of possibly variable length \(T_i\). At token position \(t\in\{1,\dots,T_i\}\), the state is \(s_{i,t}=(\mathbf{q},\,\mathbf{o}_{i,<t})\) and the action is \(a_{i,t}=o_{i,t}\), where \(\mathbf{o}_{i,<t}=(o_{i,1},\dots,o_{i,t-1})\), transitions satisfy \(\mathbb{P}(s_{i,t+1}\mid s_{i,t},a_{i,t})=1\) with \(s_{i,t+1}=(\mathbf{q},\,\mathbf{o}_{i,<t},o_{i,t})\), the process initializes at \(s_{i,1}=\mathbf{q}\), and generation terminates when the end-of-sequence symbol \(\omega\) is emitted. Each action receives a scalar reward \(R(s_{i,t},a_{i,t})\) from an automatic verifier, a learned human-preference model, or task-specific rules. Under this formulation, learning amounts to optimizing a stochastic policy \(\pi_{\theta}(a\mid s)\) to maximize the expected cumulative reward, and the deterministic, fully observable dynamics enable fine-grained analysis and explicit control of exploration metrics such as policy entropy.

\subsubsection{GRPO.}
GRPO directly computes the advantage $A_t$ using the average reward across multiple sampled outputs, thereby eliminating the need for an additional value function in PPO.

Specifically, given a prompt \(\mathbf{q} \sim P(Q)\), we sample \(G\) rollouts \(\{\mathbf{o}_i\}_{i=1}^G\) from the current policy \(\pi_{\theta_{\text{old}}}\).
At each token position \(t\) in rollout \(i\), the likelihood ratio is defined in  Eq.~\ref{eq:ration}.
\begin{equation}
  \small
\label{eq:ration}
r_{i,t}(\theta)
= \frac{\pi_{\theta}\bigl(o_{i,t}\mid \mathbf{q},\, \mathbf{o}_{i,<t}\bigr)}
       {\pi_{\theta_{\mathrm{old}}}\bigl(o_{i,t}\mid \mathbf{q},\,\mathbf{o}_{i,<t}\bigr)}
\end{equation}
The group-relative advantage \(\hat{A}_{i,t}\) is then obtained by standardizing each return \(R_i\) within the group, defined in Eq.~\ref{eq:advantage}.
\begin{equation}
  \small
\label{eq:advantage}
\hat{A}_{i,t}
= \frac{R_i \;-\;\mathrm{Mean}\bigl(\{R_j\}_{j=1}^G\bigr)}
       {\mathrm{Std}\bigl(\{R_j\}_{j=1}^G\bigr)}.
\end{equation}
And then maximizes the clipped surrogate objective by
\begin{equation}
\begin{aligned}
\label{eq:grpo}
\mathcal{J}_{\text{GRPO}}(\theta)=
&\mathbb{E}_{\mathbf{q},\{\mathbf{o}_i\}}\Bigl[
 \tfrac{1}{G}\!\sum_{i=1}^{G}\tfrac{1}{|\mathbf{o}_i|}\!\sum_{t=1}^{|\mathbf{o}_i|}
 \Bigl(\min\!\bigl[r_{i,t}(\theta)\hat{A}_{i,t},\;
  \\
  \operatorname{clip}(r_{i,t}(\theta),&1-\varepsilon,1+\varepsilon)\hat{A}_{i,t}\bigr]
 -\beta\,\mathrm{D}_{\mathrm{KL}}(\pi_{\theta}\Vert\pi_{\text{ref}})\Bigr)
\Bigr].
\end{aligned}
\end{equation}
Notably, 
\(
D_{\mathrm{KL}}[\pi_\theta\|\pi_{\mathrm{ref}}]
\)
acts as a regularization term that constrains the updated policy \(\pi_\theta\) to remain close to the reference policy \(\pi_{\mathrm{ref}}\), representing one of the earliest efforts to prevent entropy collapse and some studies claimed to be the key parameter enabling GRPO to sustain prolonged training.
\subsection{CURE}

\label{sec:method}
CURE employs a two-stage procedure. 
As shown in Fig.~\ref{fig:main}, in Stage 1, exploration is injected by dynamically reshaping the prompt distribution based on token-level uncertainty. In Stage 2, exploitation is applied by continuing training with DAPO under static initial-state sampling on a fixed corpus, consolidating Stage 1 gains into higher accuracy and overall performance.



\subsubsection{3.2.1 CURE First Stage.}
\label{sec:cure-grpo}
By sampling high-entropy ended prefixes and re-prompting, CURE's first stage explicitly injects novel yet coherent initial states, delaying premature entropy collapse and improving exploration efficiency. 
\begin{enumerate}

  \item \textbf{Initial Rollouts.}
 For each query \(\mathbf{q}\) drawn directly from the dataset, we first sample $N_1$
  trajectories from the old policy to estimate token-level uncertainty.
  \begin{equation}
   \small
    \mathcal{G}(\mathbf{q},N_1)=\{\mathbf{o}_i\}_{i=1}^{N_1}\sim\pi_{\theta_{\text{old}}}(\cdot\mid \mathbf{q}).
    \label{eq:n1}
  \end{equation}

  \item \textbf{Token-Level Entropy.}
  We compute the policy entropy at every position
  in each trajectory to detect where the model is most uncertain, which will
  guide us to unexplored-but-coherent regions of the state space.
    \begin{equation}
     \small
    \begin{aligned}
    H_{i,t} &= -\sum_{v\in\mathcal V}
      \pi_{\theta_{\text{old}}}\bigl(v\mid \mathbf{q}, \mathbf{o}_{i,<t}\bigr)
      \log \pi_{\theta_{\text{old}}}\bigl(v\mid \mathbf{q}, \mathbf{o}_{i,<t}\bigr), \\
    t &= 1,\dots,T_i.
    \end{aligned}
    \label{eq:token_entropy_full}
    \end{equation}
  \item \textbf{Top-$K$ Selection with Stochastic Choice.} We rank positions by entropy and uniformly select one index from the top-$K$ most uncertain positions to avoid a
  deterministic bias toward the single highest-entropy token.
  \begin{equation}
   \small
  \mathcal{T}^{(i)}_K
    = \operatorname{TopK}_t\!\bigl(H_{i,t}, K\bigr),\qquad
  t_i^{\star} \sim \mathrm{Uniform}\!\bigl(\mathcal{T}^{(i)}_K\bigr).
  \label{eq:t_star_full}
  \end{equation}

  \item \textbf{Frontier Prefix and Refined Prompt.} We take the prefix $\mathbf{p}_i$ up to (but not including) the sampled
  position $t_i^\star$ and prepend it to the original query, creating a refined prompt $\mathbf{q}'_i$
  that stays semantically consistent yet was unseen during prior training.
  \begin{equation}
  \small
  \mathbf{p}_i = \mathbf{o}_{i,1:t_i^\star-1},\qquad
  \mathbf{q}'_i = \mathbf{q} \,\|\, \mathbf{p}_i .
  \label{eq:refined_prompt_full}
  \end{equation}

  \item \textbf{Re-Prompting Rollouts.} Each refined prompt $q'_i$ is then fed back to the policy
  to produce $N_2$ additional trajectories, yielding a total of $N_1*N_2$
  re-prompted samples.
  \begin{equation}
    \small
  \mathcal{G}(\mathbf{q}'_i,N_2)=\{\mathbf{o}_{j}\}_{j=1}^{N_2}
  \sim \pi_{\theta_{\text{old}}}(\cdot\mid \mathbf{q}'_i).
  \label{eq:new_rollouts_full}
  \end{equation}

  \item \textbf{Group Construction.} For each dataset‐derived query \(\mathbf{q}\), we merge the original trajectory with its re-prompted counterparts to form the group \(\mathcal{G}(\mathbf{q})\). 
  This formulation underpins the computation of our GRPO-like objective across all trajectories in \(\mathcal{G}(\mathbf{q})\).
      \begin{equation}
      \begin{aligned}
    \small
    \mathcal{G}(\mathbf{q})
    &= \mathcal{G}(\mathbf{q},N_1)\;\cup\;\Bigl(\,\bigcup_{i=1}^{N_1}\mathcal{G}(\mathbf{q}'_{i},N_2)\Bigr)\\
    \bigl|\mathcal{G}(\mathbf{q})\bigr| &= N_1 + N_1*N_2.
     \end{aligned}
  \label{eq:new_rollouts_full_2}
  \end{equation}
  
\item \textbf{Batch-Level Dynamic Sampling Construction.} To improve training efficiency, we follow DAPO: at each sampling round, we discard and resample any prompts whose group of \(G\) rollouts is entirely correct or entirely incorrect, as such groups provide minimal gradient information and accelerate premature determinization.
  \item \textbf{Objective Function.} 
  We jointly optimize all trajectories in \(\mathcal{G}(\mathbf{q})\) by Eq.~\ref{eq:cure_grpo_obj}. Here $\textbf{gt}$ is the ground truth. The importance weight \(r_{i,t}(\theta)\)
is computed by Eq.~\ref{eq:ration}, and the group-relative advantage \(\hat A^{\mathrm{grp}}_{i,t}\) is computed by Eq.~\ref{eq:advantage}. 
\begin{equation}
\small
\begin{aligned}
\mathcal{J}_{\text{CURE}}(\theta)
=\mathbb{E}_{\mathbf{q}\sim P(Q)}\Biggl[\,
    \frac{1}{\sum_{o_i\in \mathcal{G}(\mathbf{q})}|\mathbf{o_i}|}
    \sum_{\mathbf{o}_i\in\mathcal{G}(\mathbf{q})}
    \sum_{t=1}^{|\mathbf{o_i}|}\\
    \Bigl(\,
      \min
      \!\bigl[
        r_{i,t}(\theta)\,
        \hat{A}_{i,t}^{\mathrm{grp}},\,
        \operatorname{clip}\bigl(r_{i,t}(\theta),\,1-\varepsilon,\,1+\varepsilon\bigr)
        \,\hat{A}_{i,t}^{\mathrm{grp}}
      \bigr]\Bigl)\Biggl]\\
\text{s.t.} \quad {0<\Big| \{\mathbf{o}\mid\texttt{is\_eq} (\mathbf{gt},\mathbf{o}),\mathbf{o}\in\mathcal{G}(\mathbf{q})\}\Big|< |\mathcal{G}(\mathbf{q})|}.
\end{aligned}
\label{eq:cure_grpo_obj}
\end{equation}
Empirically, this objective sustains higher policy entropy and improves rewards, with the KL-divergence regularization term omitted.
\end{enumerate}
\begin{algorithm}[h]
\caption{CURE Stage 1}
\label{alg:cure-stage1}
\textbf{Input}: Training set loader, rollout counts $N_1$, $N_2$, top-$K$ critical tokens, reward model $\mathcal{R}$, batch size $B$\\
\textbf{Initialize}: Parameter buffer $\mathcal{B} \leftarrow \emptyset$, step counter $g \leftarrow 0$

\begin{algorithmic}[1]
\FOR{each epoch $=1$ to $T$}
    \FOR{each mini-batch $\mathbf{q}$ in loader}
        \STATE $\mathbf{o}_1 \leftarrow$ \textsc{Generate}$(\mathbf{q}, n=N_1)$ \hfill\textcolor{gray}{// primary rollouts}
        \STATE $\{t_i^\star\} \leftarrow$ \textsc{SelectCriticalTokens}$(\mathbf{o}_1, \text{top-}K)$
        \STATE $\mathbf{q}' \leftarrow$ \textsc{ReconcatQuery}$(\mathbf{q}, \mathbf{o}_1, \{t_i^\star\})$
        \STATE $\mathbf{o}_2 \leftarrow$ \textsc{Generate}$(\mathbf{q}', n=N_2)$ \hfill\textcolor{gray}{// branch rollouts}
        \STATE $\mathcal{T} \leftarrow$ \textsc{Merge}$(\mathbf{o}_1, \mathbf{o}_2)$
        \STATE $\mathcal{T}' \leftarrow$ \textsc{Filter}$(\mathcal{T})$ \hfill\textcolor{gray}{// dynamic sampling}

        \STATE $\mathcal{B} \leftarrow \mathcal{B} \cup \mathcal{T}'$
        \IF{$|\mathcal{B}| < B$}
            \STATE \textbf{continue}
        \ENDIF
        \STATE $\mathcal{D} \leftarrow$ First $B$ samples in $\mathcal{B}$; $\mathcal{B} \leftarrow \emptyset$
        \STATE $\mathcal{L} \leftarrow$ \textsc{DAPOOptimize}$(\mathcal{D})$
        \STATE \textsc{UpdateParameters}$(\mathcal{L})$
        \STATE $g \leftarrow g + 1$
    \ENDFOR
\ENDFOR
\end{algorithmic}
\end{algorithm}
\subsubsection{3.2.2 CURE Second Stage.}




\begin{table*}[h]
\centering
\small
\setlength{\tabcolsep}{3pt}
\renewcommand{\arraystretch}{1.25}

\begin{tabular}{lccccccc}
\toprule

\textbf{Model}
& \textbf{AIME24} & \textbf{AIME25} & \textbf{AMC23} 
& \textbf{MATH500} & \textbf{Minerva} & \textbf{Olympiad} & \textbf{Avg.} \\
\midrule
Qwen-Math-7B-Base~\cite{yang2024qwen2}     & $16.6$ & $6.3$ & $52.2$ & $52.4$ & $10.7$  & $19.0$ & $26.2$ \\
Qwen-Math-7B-Instruct~\cite{yang2024qwen2} & $13.3$ & $10.0$ & $57.1$ & $84.2$ & $41.5$ & $44.4$ & $41.8$ \\
\midrule
\multicolumn{8}{c}{\textit{Previous Classical RLVR methods}}\\

\midrule
Eurus-2-7B-PRIME-Zero~\cite{cui2025process}          & $18.9$ & $11.7$ & $57.7$ & $79.8$ & $41.5$ & $48.0$ & $42.9$ \\
SimpleRL-Zero~\cite{zeng2025simplerl}      & $26.7$ & $9.3$  & $60.8$ & $77.4$ & $32.0$ & $41.5$ & $41.3$ \\
OpenReasoner-Zero~\cite{hu2025open}       & $15.4$   &    $13.4$    & $56.5$    & $80.6$   & $39.0$    & $45.9$    & $41.8$       \\
Oat-Zero~\cite{liu2025understanding}       & $28.8$ & $10.8$ & $65.2$ & $80.0$ & $42.3$ & $43.7$ & $45.1$ \\
LUFFY~\cite{yan2025learning}                 & $25.8$    & $22.3$    &  $71.7$  &  $87.0$  & $44.9$    &  $55.9$   &  $51.3$  \\
NFT~\cite{chen2025bridging} & $32.0$ & $18.3$ & $88.5$ & $83.2$ & $40.8$ & $47.3$ & $51.7$ \\
\midrule
\multicolumn{8}{c}{\textit{Previous Entropy Control methods}}\\
\midrule
KL-Cov~\cite{cui2025entropy}    & $33.4$ & $17.1$ & $77.1$ & $83.8$ & $43.0$ & $49.9$ & $50.7$ \\
Clip-Cov~\cite{cui2025entropy} & $32.4$ & $14.3$ & $81.6$ & $84.8$ & $44.5$ & $48.0$ & $50.9$ \\
\midrule
\multicolumn{8}{c}{\textit{Our Methods}}\\
\midrule
CURE First Stage  & $33.4$ & $15.3$ & $82.7$ & $82.4$ &$48.2$&$50.5$&$52.1$ \\
CURE Second Stage & $35.5$ & $18.5$ & $89.7$ & $83.4$ & $48.2$ & $50.5$ & $54.3$ \\
\bottomrule
\end{tabular}
\caption{Performance of CURE and prior RLVR methods: avg@32 on AIME24, AIME25, and AMC23; avg@1 on MATH500, Minerva, and Olympiad.}
\label{tab:main_table}
\end{table*}
Leveraging DAPO’s inherently low-entropy training dynamics, we neither decay the learning rate nor introduce an explicit KL regularizer. Instead, we train directly on all dataset-derived queries  in their original form, without any re-concatenation. In practice, we perform a 10-step warmup and then continue training to step 100 with a fixed learning rate of \(1\times10^{-6}\). As shown in Sec.~\ref{sec:experiment}, unlike DAPO baseline—where entropy continuously decreases at all temperatures under extended training—when sampling at a temperature of $1.0$ during training, our average policy entropy does not decrease. This demonstrates that our model can sustain diverse exploration at high temperatures even after continuing training. 
At the optimal evaluation temperature of $0.6$, however, this procedure yields a $29.2\%$ reduction in policy entropy and a $7.6\%$ improvement in evaluation performance. Consequently, learning becomes biased toward the high-reward behaviors discovered in Stage~1, effectively converting exploration into stable accuracy gains.

\subsubsection{3.3 Advantage}
\paragraph{Plug-and-Play with RLVR.}
CURE can be dropped into existing pipelines with minimal engineering: replace the sampler with our critical-token--guided re-concatenation routine and switch to a simple two-phase entropy schedule. Empirically, this decoupled explore-then-exploit design arrests entropy collapse, maintains policy diversity when it matters, and then selectively compresses it to harvest performance.

\paragraph{Sustained, High‑Entropy Exploration.}
By dynamically re‑prompting on high‑uncertainty prefixes and normalizing advantages across groups, CURE maintains a persistently elevated policy entropy even in late training, thereby continuously steering the model into novel but coherent regions of the state space and avoiding premature convergence to deterministic behaviors.
\paragraph{Efficient Conversion of Exploration into Performance.}
The strong positive coupling we observe between entropy and reward demonstrates that CURE’s exploration is not random noise, but directly fuels learning,yielding higher rewards per unit of entropy than DAPO, GRPO, KL-CoV, or Clip-CoV. Moreover, the two‑phase entropy schedule ensures that the diversity injected in the first stage is rapidly consolidated into accuracy gains in the second stage.

Further evidence of CURE's effectiveness is provided in the Appendix A.

\section{Experiments}
\label{sec:experiment}
We design our empirical study to answer the following research questions:
\begin{itemize}
\item \textbf{Q1: Overall Performance.} Does CURE improve mathematical reasoning compared to baseline methods?

\item \textbf{Q2: Entropy Preservation.} Can CURE sustain exploration throughout training and benefit from preventing policy-entropy collapse?

\item \textbf{Q3: Second-Stage Research.} Given that CURE generates first-stage models with significantly higher policy entropy, how can we leverage these high-entropy checkpoints to further improve reasoning performance?

\item \textbf{Q4: Ablation on Critical‑Token Strategy.}  How do alternative ways of selecting or handling critical tokens affect performance and entropy?

\end{itemize}

\subsection{Experiment Setups}
\paragraph{Training.}
 perform  fine-tuning on Qwen2.5-Math-7B-Base~\cite{yang2024qwen2} using the publicly available DAPO-Math-17K~\cite{yu2025dapo} dataset, which contains only math questions paired with integer ground-truth answers. Specifically, we omit both the KL-divergence and entropy loss terms. Rollouts are generated with a batch size of $512$, using a temperature of $1.0$ and $16$ rollouts per prompt. During policy updates, we use an update batch size of $32$. Finally, we adopt the same reward function as DAPO~\cite{yu2025dapo}, without any additional formatting reward. Further details are provided in the Appendix B.
\begin{figure*}[t]
\small
\vspace{3mm}
\centering
\includegraphics[width = \linewidth]{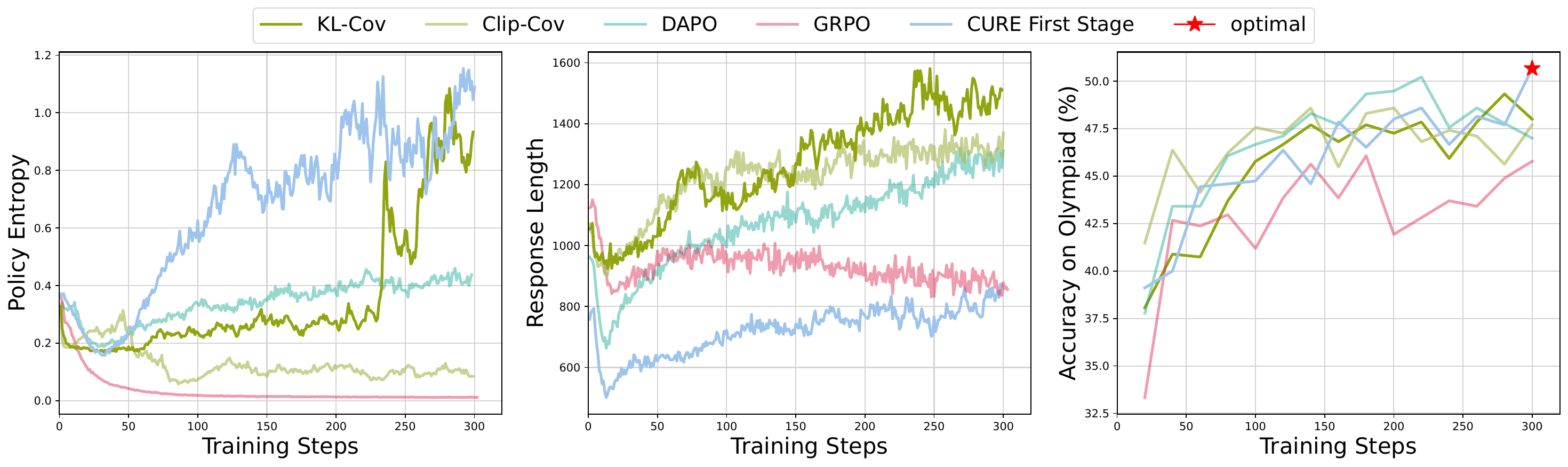}
\caption{Depicts the evolution of policy entropy and sequence length on the training set, alongside validation accuracy on AIME24, plotted over training steps for various methods.}
\label{fig:compare_three_plot}
\end{figure*}

\paragraph{Evulation.}
For evaluation, we benchmark six widely used math-reasoning datasets: AIME24~\cite{li2024numinamath}, AIME25~\cite{li2024numinamath}, AMC23~\cite{li2024numinamath} , MATH500~\cite{dataset_math}, OlympiadBench~\cite{dataset_olympiad}, and Minerva Math~\cite{dataset_minerva}. All inference model setups and sampling parameters are drawn from their official reports. Our method employs a top-p of 0.7 and a temperature of 0.6. We report avg@32 for AIME24, AIME25, and AMC23 and avg@1 for MATH-500, Minerva, and OlympiadBench. Following SimpleRL~\cite{zeng2025simplerl} and NFT~\cite{chen2025bridging}, we combine the Math-Verify~\cite{Kydlicek_Math-Verify_Math_Verification}, MathRuler-verifier~\cite{mathruler}, and SimpleRL verifier~\cite{zeng2025simplerl} for final evaluation.

\paragraph{Baselines.} We compare our method against six RLVR baselines: Eurus-2-7B-PRIME-Zero \cite{cui2025process}, SimpleRL-Zero \cite{zeng2025simplerl}, Open-Reasoner-Zero \cite{hu2025open}, Oat-Zero \cite{liu2025understanding}, LUFFY \cite{yan2025learning}, and NFT \cite{chen2025bridging}, as well as two recent entropy‐control algorithms: Clip‐Cov~\cite{cui2025entropy} and KL‐Cov~\cite{cui2025entropy}. Since no results or checkpoints were released, we used the official codebase and scripts to train the models to convergence. Detailed descriptions of all methods appear in the Appendix B.
\subsection{Overall Performance}

As shown in Tab.~\ref{tab:main_table}, our two-stage CURE framework achieves SOTA performance on diverse mathematical benchmarks with high data efficiency, outperforming both RLVR and entropy-control methods. Starting from the Qwen-7B-Math-Base model and leveraging only 17 K samples, Stage 1 of CURE achieves an average score of $52.1\%$, surpassing all other methods such as \textsc{Clip-Cov} ($50.9\%$) and \textsc{NFT} ($51.7\%$). In the second, critic-bootstrapped stage, CURE further raises the average accuracy to $54.3\%$, with particularly marked gains on AIME24 ($35.5\%$ vs.\ $26.6\%$) and AMC23 ($89.7\%$ vs.\ $52.2\%$), representing a $107\%$ improvement over the base model.
\begin{figure}[b]
\centering
\includegraphics[width=.98\columnwidth]{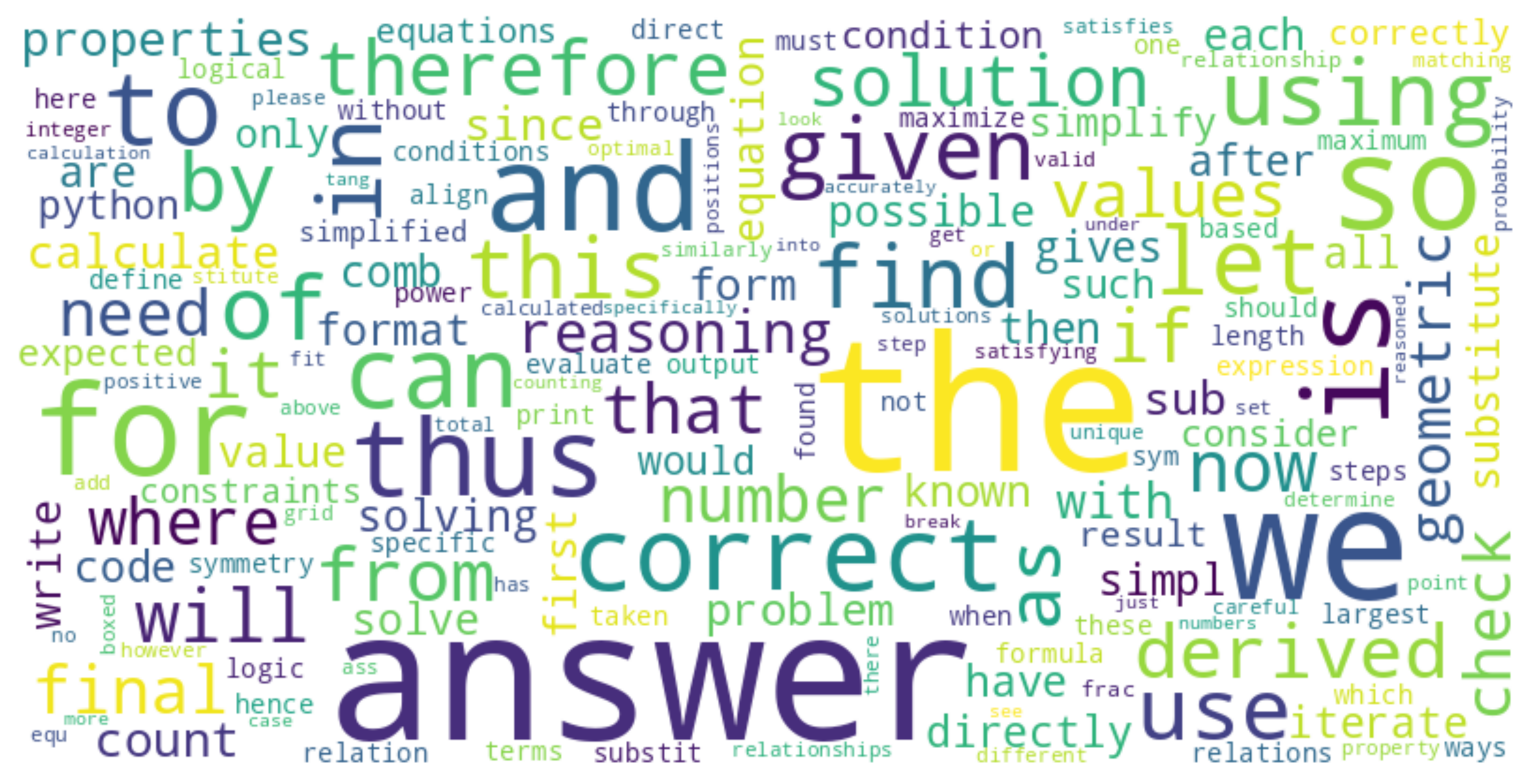} 
\caption{Word clouds generated by CURE First Stage.}
\vspace{-10pt}
\label{fig:entropy_wordclouds}
\end{figure}
\subsection{Entropy Preservation}
Fig.~\ref{fig:compare_three_plot} plots, on the training set, (i) policy entropy vs. training steps and (ii) response length vs. training steps, together with (iii) OlympiadBench accuracy vs. training steps. From the left plot (policy entropy), the CURE First Stage exhibits the steepest entropy growth, indicating stronger exploration via a more diverse policy distribution. By contrast, GRPO’s entropy collapses rapidly toward zero—signaling a loss of exploration—while DAPO and Clip-CoV plateau at moderate levels, reflecting only limited entropy regulation. As for KL-CoV, its entropy remains relatively low early on and then rises markedly in later training, approaching a higher-entropy regime. The middle plot (Response Length) shows that the CURE First Stage’s responses grow longer throughout training; however, owing to its design—where queries splice responses before key tokens—the responses generated by CURE remain relatively shorter than those of other methods even as training advances. Meanwhile, the right plot (accuracy) indicates that the CURE First Stage achieves accuracy comparable to competing methods, striking a balance between exploration (driven by entropy) and exploitation (reflected in accuracy). Notably, at the stage-one high-entropy checkpoint, our first-stage model already delivers strong results.

As illustrated in Fig.~\ref{fig:model_relase_1} and Tab.~\ref{tab:main_table}, CURE consistently outperforms other RLVR approaches—particularly entropy‐control algorithms—achieving the highest entropy across all six test sets.

\begin{figure}[h]
\centering
\includegraphics[width=\columnwidth]{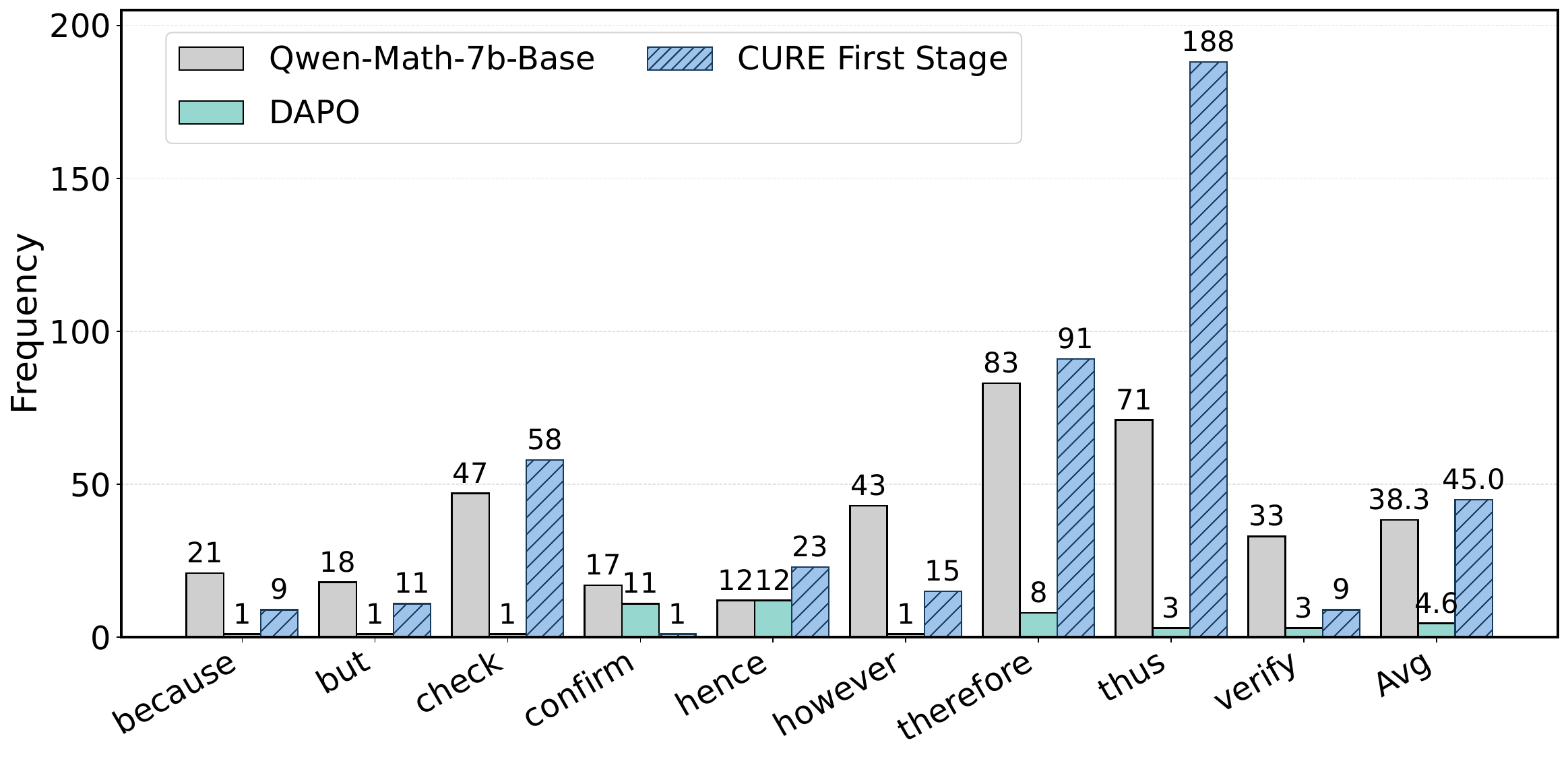} 
\caption{Probability of semantically meaningful high-entropy tokens across training paradigms in AIME24.}
\label{fig:token_freq}
\end{figure}

To further probe our high-entropy models, 
Fig.~\ref{fig:entropy_wordclouds} shows the word cloud for CURE’s first stage in AIME24, while Fig.~\ref{fig:token_freq} compares the frequencies of several key high-entropy connector tokens under GRPO, DAPO, and CURE. Consistent with Beyond the 80/20 Rule~\cite{wang2025beyond}, which highlights these tokens as markers of new reasoning pathways, we observe that CURE not only restores their baseline usage but amplifies it, whereas DAPO’s strong entropy penalty significantly suppresses them. 
After first-stage CURE training, such as “check” increases from $47$ to $58$ occurrences, “verify” surges from $33$ to $188$, and “hence” nearly doubles. This marked boost in connector usage shows that our high-entropy objective effectively promotes exploration and reinforces semantic transitions, providing a solid foundation for the subsequent continuing training stage.
\subsection{\textbf{Second-Stage Research}}
\begin{figure}[t]
\centering
\vspace{-5pt}
\includegraphics[width=.8\columnwidth]{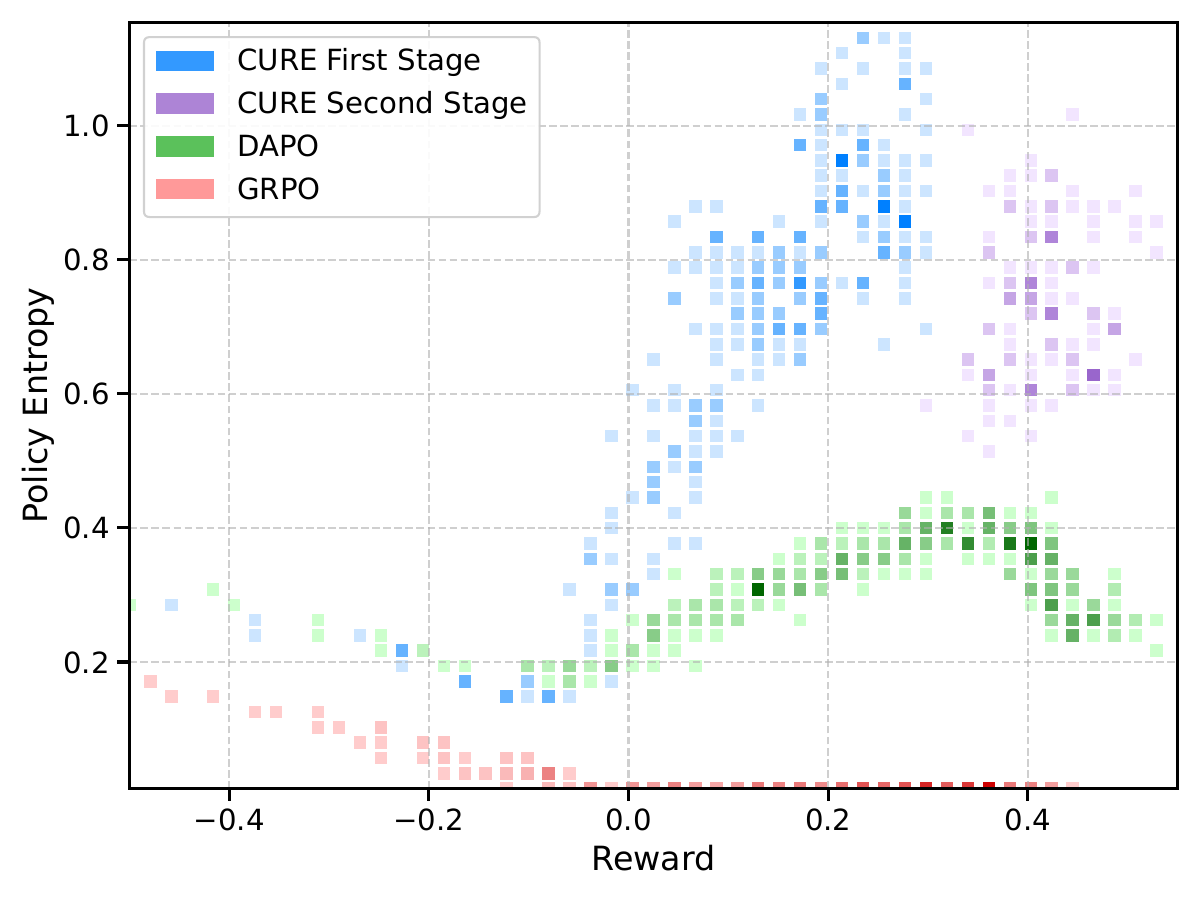} 
\caption{Scatter Plots of Policy Entropy vs. Reward for CURE , DAPO, and GRPO Methods at temperature 1.0}
\label{fig:five_entyopy}
\end{figure}
Fig.~\ref{fig:five_entyopy} depicts the evolution of training entropy and reward. As the reward increases, the policy entropy of GRPO remains very low. DAPO exhibits an initial decrease before stabilizing within a relatively low-entropy regime. 
Compared with the first stage, CURE’s second training phase exhibits markedly lower entropy overall, and the joint density of entropy and reward concentrates in the low-entropy, high-reward region—reflecting a gradual transition toward exploitation. 
Table~\ref{tab:main_table} reports performance improvements from 52.1\% to 54.3\% across six benchmark test sets, further confirming this trend.
\begin{figure}[b]
\centering
\vspace{-5mm}
\includegraphics[width=.87\columnwidth]{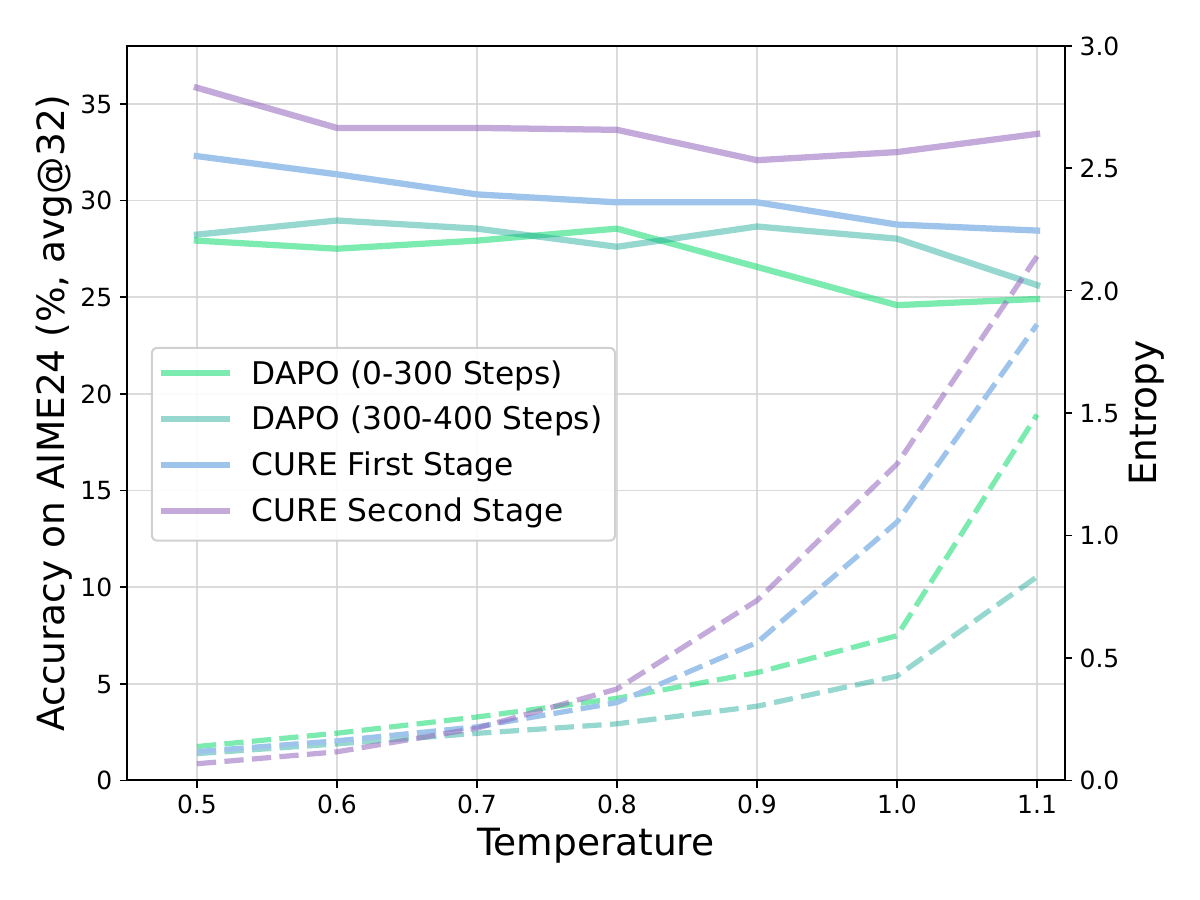}
\caption{AIME24 accuracy and entropy vs.\ temperature for DAPO and CURE at different stages}
\label{fig:temp}
\end{figure}

As shown in Fig.~\ref{fig:temp}, compared to the DAPO, CURE achieves higher scores at low temperatures ($0.6$) by maintaining lower entropy, and at high temperatures ($1.0$) facilitates broader exploration through increased entropy. By making only slight adjustments to $T$, CURE achieves a controllable shift between exploration and exploitation, striking an optimal balance that delivers a consistent $5\%$ accuracy improvement over DAPO on AIME24 across the entire temperature range.

\subsection{\textbf{Ablation on the Critical‑Token Strategy} }
\begin{table}[h]
\centering
\small
\begin{tabular}{lccccc}
\toprule
\textbf{Model} & \textbf{AIME24} & \textbf{AMC} & \textbf{Minerva} & \textbf{Oly.} & \textbf{Avg.} \\
\midrule
 Base
      & 16.6
      & 52.2
      & 10.7
      & 19.0
      & 24.6 \\
    GRPO
      & 31.0
      & 80.8
      & 40.1
      & 44.9
      & 49.2 \\
    \midrule
    DAPO
      & 28.9
      & 84.7
      & 47.4
      & 47.0
      & 52.0 \\
    CURE$_{\text{Random}}$
      & 35.7
      & 80.3
      & 41.5
      & 48.2
      & 51.4\tiny{$\downarrow0.6$} \\
    CURE$_{\text{Entropy}}$
      & 33.4
      & 82.7
      & 48.2
      & 50.5
      & 53.7\tiny{$\uparrow1.7$} \\
    \bottomrule
\end{tabular}
\caption{Comparison of performance across various strategies on benchmark math-reasoning datasets.}
\label{tab:ablation_grpo_dapo}
\end{table}

 To isolate the effect of how we choose the truncation point \(t^\star\) from Eq.~\ref{eq:t_star_full}, we compare two variants under the same training budget and hyperparameters, and include DAPO as a reference. CURE\textsubscript{Random} randomly selects a token for truncation and re-concatenation, preserving intervention frequency while ignoring token criticality. CURE\textsubscript{Entropy} selects a token from the top-$20$ highest-entropy positions, explicitly steering the model toward states of maximal distributional uncertainty.
Finally, as Tab.~\ref{tab:ablation_grpo_dapo}, CURE\textsubscript{Entropy} achieves the best average performance, $53.7\%$, outperforming DAPO by +$1.7\%$ and CURE\textsubscript{Random} by +$2.3\%$.  In contrast, in terms of raw performance, CURE\textsubscript{Random} remains close to DAPO ($51.4\%$ vs.\ $52.0\%$, -$0.6\%$). When we examine the training‐set entropy, CURE\textsubscript{Random} shows only a $73\%$ increase—far below the $137\%$ gain of CURE\textsubscript{Entropy} both measured relative to the pre-training baseline.
These results indicate that appending unguided random prefixes is ineffective, the intervention location is pivotal. Prioritizing high-entropy tokens offers a principled and effective criterion for selecting critical intervention points.

\section{Conclusion}
We identify entropy collapse induced by static initial‑state sampling as a key bottleneck for RLVR‑based reasoning in LLMs and present CURE, 
a two‑stage framework that couples targeted exploration with stable exploitation. 
By re‑generating at high‑entropy critical tokens and re‑concatenating coherent branches, 
the first stage injects structured diversity without relying on fragile global entropy bonuses, the second stage then continues training with DAPO’s static starts to consolidate gains in a controlled manner. 
On Qwen-2.5-Math-7B, 
CURE eventually outperforms other RLVR methods by an average of 5\% across six math benchmarks, 
while sustaining elevated policy entropy thereby establishing a SOTA performance. 
Our experiments confirm that both critical‑token branching and the subsequent continuing training are necessary to prevent premature determinism and to translate exploration into accuracy.

Looking forward, CURE’s data‑generation intervention is orthogonal to reward shaping and KL‑based regularization, suggesting broad applicability across models, tasks, and verification protocols. Future work includes adaptive scheduling of the two stages, principled detection of critical tokens beyond math (e.g., code and multimodal reasoning), theoretical analysis of entropy dynamics under branching, and systems‑level optimizations for efficient batched regeneration. We believe this line of inquiry offers a practical recipe for sustained reasoning gains in RLVR: maintain exploration where it matters, then exploit deliberately.
\bibliography{aaai2026,luppy}

\begin{thebibliography}{36}
\providecommand{\natexlab}[1]{#1}

\bibitem[{Burda et~al.(2018)Burda, Edwards, Storkey, and Klimov}]{burda2018exploration}
Burda, Y.; Edwards, H.; Storkey, A.; and Klimov, O. 2018.
\newblock Exploration by random network distillation.
\newblock \emph{arXiv preprint arXiv:1810.12894}.

\bibitem[{Chen et~al.(2025)Chen, Zheng, Zhang, Cui, Cui, Ye, Lin, Liu, Zhu, and Wang}]{chen2025bridging}
Chen, H.; Zheng, K.; Zhang, Q.; Cui, G.; Cui, Y.; Ye, H.; Lin, T.-Y.; Liu, M.-Y.; Zhu, J.; and Wang, H. 2025.
\newblock Bridging supervised learning and reinforcement learning in math reasoning.
\newblock \emph{arXiv preprint arXiv:2505.18116}.

\bibitem[{Cheng et~al.(2025)Cheng, Huang, Zhu, Dai, Zhao, Zhang, and Wei}]{cheng2025reasoning}
Cheng, D.; Huang, S.; Zhu, X.; Dai, B.; Zhao, W.~X.; Zhang, Z.; and Wei, F. 2025.
\newblock Reasoning with exploration: An entropy perspective.
\newblock \emph{arXiv preprint arXiv:2506.14758}.

\bibitem[{Christiano et~al.(2017)Christiano, Leike, Brown, Martic, Legg, and Amodei}]{christiano2017deep}
Christiano, P.~F.; Leike, J.; Brown, T.; Martic, M.; Legg, S.; and Amodei, D. 2017.
\newblock Deep reinforcement learning from human preferences.
\newblock \emph{Advances in neural information processing systems}, 30.

\bibitem[{Cui et~al.(2024)Cui, Yuan, Ding, Yao, He, Zhu, Ni, Xie, Xie, Lin, Liu, and Sun}]{Cui2023ULTRAFEEDBACKBL}
Cui, G.; Yuan, L.; Ding, N.; Yao, G.; He, B.; Zhu, W.; Ni, Y.; Xie, G.; Xie, R.; Lin, Y.; Liu, Z.; and Sun, M. 2024.
\newblock ULTRAFEEDBACK: Boosting Language Models with Scaled AI Feedback.
\newblock In \emph{ICML}.

\bibitem[{Cui et~al.(2025{\natexlab{a}})Cui, Yuan, Wang, Wang, Li, He, Fan, Yu, Xu, Chen et~al.}]{cui2025process}
Cui, G.; Yuan, L.; Wang, Z.; Wang, H.; Li, W.; He, B.; Fan, Y.; Yu, T.; Xu, Q.; Chen, W.; et~al. 2025{\natexlab{a}}.
\newblock Process reinforcement through implicit rewards.
\newblock \emph{arXiv preprint arXiv:2502.01456}.

\bibitem[{Cui et~al.(2025{\natexlab{b}})Cui, Zhang, Chen, Yuan, Wang, Zuo, Li, Fan, Chen, Chen et~al.}]{cui2025entropy}
Cui, G.; Zhang, Y.; Chen, J.; Yuan, L.; Wang, Z.; Zuo, Y.; Li, H.; Fan, Y.; Chen, H.; Chen, W.; et~al. 2025{\natexlab{b}}.
\newblock The entropy mechanism of reinforcement learning for reasoning language models.
\newblock \emph{arXiv preprint arXiv:2505.22617}.

\bibitem[{DeepSeek-AI et~al.(2025)DeepSeek-AI, Guo, Yang, Zhang, Song, Zhang, Xu, Zhu, Ma, Wang, Bi, Zhang, Yu, Wu, Wu, Gou, Shao, Li, Gao, Liu, Xue, Wang, Wu, Feng, Lu, Zhao, Deng, Zhang, Ruan, Dai, Chen, Ji, Li, Lin, Dai, Luo, Hao, Chen, Li, Zhang, Bao, Xu, Wang, Ding, Xin, Gao, Qu, Li, Guo, Li, Wang, Chen, Yuan, Qiu, Li, Cai, Ni, Liang, Chen, Dong, Hu, Gao, Guan, Huang, Yu, Wang, Zhang, Zhao, Wang, Zhang, Xu, Xia, Zhang, Zhang, Tang, Li, Wang, Li, Tian, Huang, Zhang, Wang, Chen, Du, Ge, Zhang, Pan, Wang, Chen, Jin, Chen, Lu, Zhou, Chen, Ye, Wang, Yu, Zhou, Pan, Li, Zhou, Wu, Ye, Yun, Pei, Sun, Wang, Zeng, Zhao, Liu, Liang, Gao, Yu, Zhang, Xiao, An, Liu, Wang, Chen, Nie, Cheng, Liu, Xie, Liu, Yang, Li, Su, Lin, Li, Jin, Shen, Chen, Sun, Wang, Song, Zhou, Wang, Shan, Li, Wang, Wei, Zhang, Xu, Li, Zhao, Sun, Wang, Yu, Zhang, Shi, Xiong, He, Piao, Wang, Tan, Ma, Liu, Guo, Ou, Wang, Gong, Zou, He, Xiong, Luo, You, Liu, Zhou, Zhu, Xu, Huang, Li, Zheng, Zhu, Ma, Tang, Zha, Yan, Ren, Ren, Sha, Fu, Xu, Xie, Zhang,
  Hao, Ma, Yan, Wu, Gu, Zhu, Liu, Li, Xie, Song, Pan, Huang, Xu, Zhang, and Zhang}]{deepseekai2025deepseekr1incentivizingreasoningcapability}
DeepSeek-AI; Guo, D.; Yang, D.; Zhang, H.; Song, J.; Zhang, R.; Xu, R.; Zhu, Q.; Ma, S.; Wang, P.; Bi, X.; Zhang, X.; Yu, X.; Wu, Y.; Wu, Z.~F.; Gou, Z.; Shao, Z.; Li, Z.; Gao, Z.; Liu, A.; Xue, B.; Wang, B.; Wu, B.; Feng, B.; Lu, C.; Zhao, C.; Deng, C.; Zhang, C.; Ruan, C.; Dai, D.; Chen, D.; Ji, D.; Li, E.; Lin, F.; Dai, F.; Luo, F.; Hao, G.; Chen, G.; Li, G.; Zhang, H.; Bao, H.; Xu, H.; Wang, H.; Ding, H.; Xin, H.; Gao, H.; Qu, H.; Li, H.; Guo, J.; Li, J.; Wang, J.; Chen, J.; Yuan, J.; Qiu, J.; Li, J.; Cai, J.~L.; Ni, J.; Liang, J.; Chen, J.; Dong, K.; Hu, K.; Gao, K.; Guan, K.; Huang, K.; Yu, K.; Wang, L.; Zhang, L.; Zhao, L.; Wang, L.; Zhang, L.; Xu, L.; Xia, L.; Zhang, M.; Zhang, M.; Tang, M.; Li, M.; Wang, M.; Li, M.; Tian, N.; Huang, P.; Zhang, P.; Wang, Q.; Chen, Q.; Du, Q.; Ge, R.; Zhang, R.; Pan, R.; Wang, R.; Chen, R.~J.; Jin, R.~L.; Chen, R.; Lu, S.; Zhou, S.; Chen, S.; Ye, S.; Wang, S.; Yu, S.; Zhou, S.; Pan, S.; Li, S.~S.; Zhou, S.; Wu, S.; Ye, S.; Yun, T.; Pei, T.; Sun, T.; Wang, T.; Zeng, W.;
  Zhao, W.; Liu, W.; Liang, W.; Gao, W.; Yu, W.; Zhang, W.; Xiao, W.~L.; An, W.; Liu, X.; Wang, X.; Chen, X.; Nie, X.; Cheng, X.; Liu, X.; Xie, X.; Liu, X.; Yang, X.; Li, X.; Su, X.; Lin, X.; Li, X.~Q.; Jin, X.; Shen, X.; Chen, X.; Sun, X.; Wang, X.; Song, X.; Zhou, X.; Wang, X.; Shan, X.; Li, Y.~K.; Wang, Y.~Q.; Wei, Y.~X.; Zhang, Y.; Xu, Y.; Li, Y.; Zhao, Y.; Sun, Y.; Wang, Y.; Yu, Y.; Zhang, Y.; Shi, Y.; Xiong, Y.; He, Y.; Piao, Y.; Wang, Y.; Tan, Y.; Ma, Y.; Liu, Y.; Guo, Y.; Ou, Y.; Wang, Y.; Gong, Y.; Zou, Y.; He, Y.; Xiong, Y.; Luo, Y.; You, Y.; Liu, Y.; Zhou, Y.; Zhu, Y.~X.; Xu, Y.; Huang, Y.; Li, Y.; Zheng, Y.; Zhu, Y.; Ma, Y.; Tang, Y.; Zha, Y.; Yan, Y.; Ren, Z.~Z.; Ren, Z.; Sha, Z.; Fu, Z.; Xu, Z.; Xie, Z.; Zhang, Z.; Hao, Z.; Ma, Z.; Yan, Z.; Wu, Z.; Gu, Z.; Zhu, Z.; Liu, Z.; Li, Z.; Xie, Z.; Song, Z.; Pan, Z.; Huang, Z.; Xu, Z.; Zhang, Z.; and Zhang, Z. 2025.
\newblock DeepSeek-R1: Incentivizing Reasoning Capability in LLMs via Reinforcement Learning.
\newblock arXiv:2501.12948.

\bibitem[{Ecoffet et~al.(2019)Ecoffet, Huizinga, Lehman, Stanley, and Clune}]{ecoffet2019go}
Ecoffet, A.; Huizinga, J.; Lehman, J.; Stanley, K.~O.; and Clune, J. 2019.
\newblock Go-explore: a new approach for hard-exploration problems.
\newblock \emph{arXiv preprint arXiv:1901.10995}.

\bibitem[{He et~al.(2024)He, Luo, Bai, Hu, Thai, Shen, Hu, Han, Huang, Zhang et~al.}]{dataset_olympiad}
He, C.; Luo, R.; Bai, Y.; Hu, S.; Thai, Z.; Shen, J.; Hu, J.; Han, X.; Huang, Y.; Zhang, Y.; et~al. 2024.
\newblock OlympiadBench: A Challenging Benchmark for Promoting AGI with Olympiad-Level Bilingual Multimodal Scientific Problems.
\newblock In \emph{Proceedings of the 62nd Annual Meeting of the Association for Computational Linguistics (Volume 1: Long Papers)}, 3828--3850.

\bibitem[{He et~al.(2025)He, Liu, Liu, Yan, Wang, Cheng, Zhang, Zhang, Xu, Shen et~al.}]{he2025skywork}
He, J.; Liu, J.; Liu, C.~Y.; Yan, R.; Wang, C.; Cheng, P.; Zhang, X.; Zhang, F.; Xu, J.; Shen, W.; et~al. 2025.
\newblock Skywork open reasoner 1 technical report.
\newblock \emph{arXiv preprint arXiv:2505.22312}.

\bibitem[{Hendrycks et~al.(2021)Hendrycks, Burns, Kadavath, Arora, Basart, Tang, Song, and Steinhardt}]{dataset_math}
Hendrycks, D.; Burns, C.; Kadavath, S.; Arora, A.; Basart, S.; Tang, E.; Song, D.; and Steinhardt, J. 2021.
\newblock Measuring mathematical problem solving with the math dataset.
\newblock \emph{arXiv preprint arXiv:2103.03874}.

\bibitem[{hiyouga(2025)}]{mathruler}
hiyouga. 2025.
\newblock MathRuler.
\newblock \url{https://github.com/hiyouga/MathRuler}.

\bibitem[{Hu et~al.(2025)Hu, Zhang, Han, Jiang, Zhang, and Shum}]{hu2025open}
Hu, J.; Zhang, Y.; Han, Q.; Jiang, D.; Zhang, X.; and Shum, H.-Y. 2025.
\newblock Open-reasoner-zero: An open source approach to scaling up reinforcement learning on the base model.
\newblock \emph{arXiv preprint arXiv:2503.24290}.

\bibitem[{Jaech et~al.(2024)Jaech, Kalai, Lerer, Richardson, El-Kishky, Low, Helyar, Madry, Beutel, Carney et~al.}]{jaech2024openai}
Jaech, A.; Kalai, A.; Lerer, A.; Richardson, A.; El-Kishky, A.; Low, A.; Helyar, A.; Madry, A.; Beutel, A.; Carney, A.; et~al. 2024.
\newblock Openai o1 system card.
\newblock \emph{arXiv preprint arXiv:2412.16720}.

\bibitem[{Kydlíček()}]{Kydlicek_Math-Verify_Math_Verification}
Kydlíček, H. ????
\newblock {Math-Verify: Math Verification Library}.

\bibitem[{Ladosz et~al.(2022)Ladosz, Weng, Kim, and Oh}]{ladosz2022exploration}
Ladosz, P.; Weng, L.; Kim, M.; and Oh, H. 2022.
\newblock Exploration in deep reinforcement learning: A survey.
\newblock \emph{Information Fusion}, 85: 1--22.

\bibitem[{Lewkowycz et~al.(2022)Lewkowycz, Andreassen, Dohan, Dyer, Michalewski, Ramasesh, Slone, Anil, Schlag, Gutman-Solo et~al.}]{dataset_minerva}
Lewkowycz, A.; Andreassen, A.; Dohan, D.; Dyer, E.; Michalewski, H.; Ramasesh, V.; Slone, A.; Anil, C.; Schlag, I.; Gutman-Solo, T.; et~al. 2022.
\newblock Solving quantitative reasoning problems with language models.
\newblock \emph{Advances in Neural Information Processing Systems}, 35: 3843--3857.

\bibitem[{Li et~al.(2024)Li, Beeching, Tunstall, Lipkin, Soletskyi, Huang, Rasul, Yu, Jiang, Shen et~al.}]{li2024numinamath}
Li, J.; Beeching, E.; Tunstall, L.; Lipkin, B.; Soletskyi, R.; Huang, S.; Rasul, K.; Yu, L.; Jiang, A.~Q.; Shen, Z.; et~al. 2024.
\newblock Numinamath: The largest public dataset in AI4Maths with 860k pairs of competition math problems and solutions.
\newblock Hugging Face repository, 13:9.

\bibitem[{Liu et~al.(2025{\natexlab{a}})Liu, Diao, Lu, Hu, Dong, Choi, Kautz, and Dong}]{liu2025prorl}
Liu, M.; Diao, S.; Lu, X.; Hu, J.; Dong, X.; Choi, Y.; Kautz, J.; and Dong, Y. 2025{\natexlab{a}}.
\newblock Prorl: Prolonged reinforcement learning expands reasoning boundaries in large language models.
\newblock \emph{arXiv preprint arXiv:2505.24864}.

\bibitem[{Liu et~al.(2025{\natexlab{b}})Liu, Chen, Li, Qi, Pang, Du, Lee, and Lin}]{liu2025understanding}
Liu, Z.; Chen, C.; Li, W.; Qi, P.; Pang, T.; Du, C.; Lee, W.~S.; and Lin, M. 2025{\natexlab{b}}.
\newblock Understanding r1-zero-like training: A critical perspective.
\newblock \emph{arXiv preprint arXiv:2503.20783}.

\bibitem[{Liu et~al.(2024)Liu, Chen, Shoeybi, Catanzaro, and Ping}]{liu2024acemath}
Liu, Z.; Chen, Y.; Shoeybi, M.; Catanzaro, B.; and Ping, W. 2024.
\newblock AceMath: Advancing Frontier Math Reasoning with Post-Training and Reward Modeling.
\newblock \emph{arXiv preprint arXiv:2412.15084}.

\bibitem[{Ouyang et~al.(2022)Ouyang, Wu, Jiang, Almeida, Wainwright, Mishkin, Zhang, Agarwal, Slama, Ray et~al.}]{ouyang2022training}
Ouyang, L.; Wu, J.; Jiang, X.; Almeida, D.; Wainwright, C.; Mishkin, P.; Zhang, C.; Agarwal, S.; Slama, K.; Ray, A.; et~al. 2022.
\newblock Training language models to follow instructions with human feedback.
\newblock \emph{Advances in neural information processing systems}, 35: 27730--27744.

\bibitem[{Rafailov et~al.(2023)Rafailov, Sharma, Mitchell, Manning, Ermon, and Finn}]{rafailov2023direct}
Rafailov, R.; Sharma, A.; Mitchell, E.; Manning, C.~D.; Ermon, S.; and Finn, C. 2023.
\newblock Direct preference optimization: Your language model is secretly a reward model.
\newblock \emph{Advances in neural information processing systems}, 36: 53728--53741.

\bibitem[{Rein et~al.(2024)Rein, Hou, Stickland, Petty, Pang, Dirani, Michael, and Bowman}]{rein2024gpqa}
Rein, D.; Hou, B.~L.; Stickland, A.~C.; Petty, J.; Pang, R.~Y.; Dirani, J.; Michael, J.; and Bowman, S.~R. 2024.
\newblock Gpqa: A graduate-level google-proof q\&a benchmark.
\newblock In \emph{First Conference on Language Modeling}.

\bibitem[{Tarvainen and Valpola(2017)}]{tarvainen2017mean}
Tarvainen, A.; and Valpola, H. 2017.
\newblock Mean teachers are better role models: Weight-averaged consistency targets improve semi-supervised deep learning results.
\newblock \emph{Advances in neural information processing systems}, 30.

\bibitem[{Team et~al.(2025)Team, Du, Gao, Xing, Jiang, Chen, Li, Xiao, Du, Liao et~al.}]{team2025kimi}
Team, K.; Du, A.; Gao, B.; Xing, B.; Jiang, C.; Chen, C.; Li, C.; Xiao, C.; Du, C.; Liao, C.; et~al. 2025.
\newblock Kimi k1. 5: Scaling Reinforcement Learning with LLMs.
\newblock \emph{arXiv preprint arXiv:2501.12599}.

\bibitem[{Wang et~al.(2025)Wang, Yu, Gao, Zheng, Liu, Lu, Dang, Chen, Yang, Zhang et~al.}]{wang2025beyond}
Wang, S.; Yu, L.; Gao, C.; Zheng, C.; Liu, S.; Lu, R.; Dang, K.; Chen, X.; Yang, J.; Zhang, Z.; et~al. 2025.
\newblock Beyond the 80/20 rule: High-entropy minority tokens drive effective reinforcement learning for llm reasoning.
\newblock \emph{arXiv preprint arXiv:2506.01939}.

\bibitem[{Wei et~al.(2024)Wei, Wang, Liu, Ding, and Zhang}]{wei2024magicoder}
Wei, Y.; Wang, Z.; Liu, J.; Ding, Y.; and Zhang, L. 2024.
\newblock Magicoder: Empowering code generation with oss-instruct.
\newblock In \emph{Forty-first International Conference on Machine Learning}.

\bibitem[{Yan et~al.(2025)Yan, Li, Hu, Wang, Cui, Qu, Cheng, and Zhang}]{yan2025learning}
Yan, J.; Li, Y.; Hu, Z.; Wang, Z.; Cui, G.; Qu, X.; Cheng, Y.; and Zhang, Y. 2025.
\newblock Learning to reason under off-policy guidance.
\newblock \emph{arXiv preprint arXiv:2504.14945}.

\bibitem[{Yang et~al.(2024)Yang, Zhang, Hui, Gao, Yu, Li, Liu, Tu, Zhou, Lin et~al.}]{yang2024qwen2}
Yang, A.; Zhang, B.; Hui, B.; Gao, B.; Yu, B.; Li, C.; Liu, D.; Tu, J.; Zhou, J.; Lin, J.; et~al. 2024.
\newblock Qwen2. 5-math technical report: Toward mathematical expert model via self-improvement.
\newblock \emph{arXiv preprint arXiv:2409.12122}.

\bibitem[{Yu et~al.(2025)Yu, Zhang, Zhu, Yuan, Zuo, Yue, Dai, Fan, Liu, Liu et~al.}]{yu2025dapo}
Yu, Q.; Zhang, Z.; Zhu, R.; Yuan, Y.; Zuo, X.; Yue, Y.; Dai, W.; Fan, T.; Liu, G.; Liu, L.; et~al. 2025.
\newblock Dapo: An open-source llm reinforcement learning system at scale.
\newblock \emph{arXiv preprint arXiv:2503.14476}.

\bibitem[{Yuan et~al.(2024)Yuan, Cui, Wang, Ding, Wang, Deng, Shan, Chen, Xie, Lin, Liu, Zhou, Peng, Liu, and Sun}]{Yuan2024AdvancingLR}
Yuan, L.; Cui, G.; Wang, H.; Ding, N.; Wang, X.; Deng, J.; Shan, B.; Chen, H.; Xie, R.; Lin, Y.; Liu, Z.; Zhou, B.; Peng, H.; Liu, Z.; and Sun, M. 2024.
\newblock Advancing LLM Reasoning Generalists with Preference Trees.
\newblock \emph{ArXiv}.

\bibitem[{Zeng et~al.(2025)Zeng, Huang, Liu, He, Liu, Ma, and He}]{zeng2025simplerl}
Zeng, W.; Huang, Y.; Liu, W.; He, K.; Liu, Q.; Ma, Z.; and He, J. 2025.
\newblock 7B Model and 8K Examples: Emerging Reasoning with Reinforcement Learning is Both Effective and Efficient.
\newblock \url{https://hkust-nlp.notion.site/simplerl-reason}.
\newblock Notion Blog.

\bibitem[{Zhang et~al.(2025)Zhang, Wang, Cheng, Zhuang, Lin, Zhang, Wang, Cui, Wang, Peng et~al.}]{zhang2025srpo}
Zhang, X.; Wang, J.; Cheng, Z.; Zhuang, W.; Lin, Z.; Zhang, M.; Wang, S.; Cui, Y.; Wang, C.; Peng, J.; et~al. 2025.
\newblock Srpo: A cross-domain implementation of large-scale reinforcement learning on llm.
\newblock \emph{arXiv preprint arXiv:2504.14286}.

\bibitem[{Zhou et~al.(2025)Zhou, Li, Wang, Cheng, Zhou, and Hsieh}]{zhou2025r1}
Zhou, H.; Li, X.; Wang, R.; Cheng, M.; Zhou, T.; and Hsieh, C.-J. 2025.
\newblock R1-Zero's" Aha Moment" in Visual Reasoning on a 2B Non-SFT Model.
\newblock \emph{arXiv preprint arXiv:2503.05132}.

\end{thebibliography}

\clearpage
\appendix
\begin{center}
    \fontsize{15pt}{\baselineskip}\selectfont
    \textbf{Appendix}
    \vspace{0.4cm}
\end{center}

\section{Why CURE can prevent entropy collapse?}
\paragraph{From the perspective of traditional RL.}
A common line of work in exploration treats the (state, policy) uncertainty via the policy entropy
\(\!H(\pi(\cdot\,|\,s))\!\) and adds either an explicit entropy bonus or a state exploration bonus to the return:
\begin{equation}
\begin{aligned}
J(\theta) \;=\; \mathbb{E}_{\pi_\theta}\!\big[\sum_{t=0}^\infty \gamma^t \big(r(s_t,a_t)
+ \beta\, H(\pi_\theta(\cdot\,|\,s_t)) + b(s_t)\big)\big],
\end{aligned}
\label{eq:regularized-objective}
\end{equation}
where \(b(s)\) increases visitation of uncertain or under-explored states~\cite{ladosz2022exploration,ecoffet2019go,burda2018exploration}. Building on this insight, we propose to improve exploration by directly exposing an LLM-based agent to unfamiliar states, thereby steering data collection toward regions of high uncertainty and increasing exploration efficiency.
\paragraph{Operational recap.}
We first draw \(N_1\) initial rollouts \(\mathcal{G}(\mathbf{q},N_1)=\{\mathbf{o}_i\}_{i=1}^{N_1}\sim\pi_{\theta_{\text{old}}}(\cdot\mid\mathbf{q})\). For each \(\mathbf{o}_i\), we identify a high-entropy critical position \(t_i^{\star}\) (Sec.~\ref{sec:cure-grpo}, Eqs.~\ref{eq:token_entropy_full}–\ref{eq:t_star_full}), extract the frontier prefix \(\mathbf{p}_i=\mathbf{o}_{i,1:t_i^{\star}-1}\), and construct the refined prompt \(\mathbf{q}'_i=\mathbf{q}\,\|\,\mathbf{p}_i\) (Eq.~\ref{eq:refined_prompt_full}). From each \(\mathbf{q}'_i\) we sample \(N_2\) additional trajectories \(\mathcal{G}(\mathbf{q}'_i,N_2)\) (Eq.~\ref{eq:new_rollouts_full}), then group all rollouts as
\begin{equation}
    \begin{aligned}
\small
\mathcal{G}(\mathbf{q})
&=\mathcal{G}(\mathbf{q},N_1)\;\cup\;\Bigl(\,\bigcup_{i=1}^{N_1}\mathcal{G}(\mathbf{q}'_i,N_2)\Bigr),
\quad
\\&\bigl|\mathcal{G}(\mathbf{q})\bigr|=N_1+N_1*N_2
    \end{aligned}
\end{equation}

(Eq.~\ref{eq:new_rollouts_full_2}). We finally optimize the CURE objective in Eq.~\ref{eq:cure_grpo_obj} with group-relative advantages, which empirically sustains higher policy entropy and improves rewards during exploration.

\paragraph{Why the re-prompted state is ``unfamiliar'': no gradient flows through the injected prefix.}
The policy is \emph{not} trained to traverse the path \(\mathbf{q}\to\mathbf{q}'_i=\mathbf{q}\,\|\,\mathbf{p}_i\). The prefix \(\mathbf{p}_i\) is treated as an exogenous intervention on the initial state rather than as an action sequence to be reinforced. Formalizing this with a stop–gradient operator \(\operatorname{sg}(\cdot)\),
\[
\small
\mathbf{q}'_i \;=\; \mathbf{q}\,\|\,\operatorname{sg}(\mathbf{p}_i),
\qquad
\nabla_{\theta}\,\mathbf{q}'_i \;=\; \mathbf{0},
\]
so the policy gradient contains no term that explains or rewards how to ``arrive'' at \(\mathbf{q}'_i\) from \(\mathbf{q}\). Instead, for the \(N_1 N_2\) rollouts that begin at the critical-token re-prompted state, the update includes likelihood factors only for tokens generated after the (re-)prompt is instantiated. By contrast, although the \(N_1\) original rollouts initiated from \(\mathbf{q}\) can carry gradients along trajectories that move from \(\mathbf{q}\) toward \(\mathbf{q}'_i\), they constitute only a small fraction of the batch and therefore contribute comparatively little to the overall update:
\begin{equation}
\begin{aligned}
\small
\nabla_{\theta}\,\mathcal{J}_{\text{CURE}}(\theta)
&=\mathbb{E}_{\mathbf{q}\sim P(Q)}\!\Biggl[
\frac{1}{\sum_{\mathbf{o}_i\in\mathcal{G}(\mathbf{q})}|\mathbf{o}_i|}
\\&\sum_{\mathbf{o}_i\in\mathcal{G}(\mathbf{q})}
\sum_{t=1}^{|\mathbf{o}_i|}
\nabla_{\theta}\, r_{i,t}(\theta)\,\hat A^{\mathrm{grp}}_{i,t}
\;
\Biggr], \label{eq:cure_grad_main}
\end{aligned}
\end{equation}

where \(s_{i,t}=\bigl(\tilde{\mathbf{q}}_{\mathbf{o}_i},\,\mathbf{o}_{i,<t}\bigr)\) with \(\tilde{\mathbf{q}}_{\mathbf{o}_i}\in\{\mathbf{q}\}\cup\{\mathbf{q}'_i\}_{i=1}^{N_1}\), \(r_{i,t}(\theta)\) is the importance weight in Eq.~\ref{eq:ration}, and \(\hat A^{\mathrm{grp}}_{i,t}\) is the group-relative advantage in Eq.~\ref{eq:advantage}. Crucially, there are no log-probability (hence no gradient) terms for tokens inside \(\mathbf{p}_i\). The mapping \(\mathbf{q}\mapsto\mathbf{q}'_i\) is therefore never reinforced, rendering \(\mathbf{q}'_i\) a genuinely \emph{novel} initial state that helps delay premature entropy collapse. 

\section{Additional Implementation Details}
\subsection{Baselines}

\subsubsection{Details of Baselines}
\begin{itemize}
  \item \textbf{Eurus-2-7B-PRIME-Zero}~\cite{cui2025process} is a reinforcement learning method for large language models that enables online updates of process reward models (PRMs) without requiring explicit process-level annotations. It leverages implicit process rewards derived from policy rollouts and outcome-level labels, removing the need for a separate reward model training phase. PRIME is compatible with various advantage functions and is designed to reduce reliance on costly supervision in multi-step reasoning tasks.
  \item \textbf{SimpleRL-Zero}~\cite{zeng2025simplerl}is a rule-based reinforcement-learning recipe for math reasoning Zero RL Training.
DeepSeek-R1 demonstrates that long chain-of-thought (CoT) reasoning can emerge from a simple reinforcement learning framework with rule-based rewards, starting directly from base models—a setting referred to as zero RL training. Recent work extends this paradigm to ten diverse base models, including LLama3-8B, Mistral-7B/24B, DeepSeek-Math-7B, and Qwen2.5 models from 0.5B to 32B. Key strategies include adjusting format rewards and controlling input difficulty to guide reasoning development during training.
  \item \textbf{Open-Reasoner-Zero}~\cite{hu2025open} is an open-source implementation of large-scale reasoning-oriented reinforcement learning directly on base models. ORZ adopts a minimalist approach using vanilla PPO with GAE and rule-based rewards, without KL regularization. It maintains scalability and training simplicity by eliminating auxiliary components. ORZ further incorporates a learned critic that penalizes repetitive patterns, enabling more stable advantage estimation.
  \item \textbf{Oat-Zero}~\cite{liu2025understanding}, analyzes two key components of the R1-Zero paradigm: base models and reinforcement learning algorithms. It examines how pretraining characteristics influence RL dynamics by comparing various base models, including DeepSeek-V3-Base and Qwen2.5 series. To address the optimization bias in Group Relative Policy Optimization (GRPO) that inflates response length, it introduces Dr. GRPO, an unbiased optimization method designed to improve token efficiency. Based on these analyses, it proposes a simplified R1-Zero training recipe.
  \item \textbf{LUFFY}~\cite{yan2025learning} is an off-policy RLVR framework that augments on-policy learning with external reasoning traces, allowing models to acquire abilities beyond their own outputs. It mixes off-policy demonstrations with on-policy rollouts, combining Mixed-Policy GRPO whose convergence rate is theoretically guaranteed with policy shaping via regularized importance sampling to balance imitation and exploration while avoiding brittle, surface-level mimicry. It succeeds at training weak models where on-policy RLVR fails. The framework points toward scalable, data-efficient RLVR that
leverages broad off-policy guidance while preserving exploration.
\begin{table}[bh]
\small
\begin{tabular}{lcc}
\toprule
\textbf{Settings} & \textbf{First Stage} & \textbf{Second Stage} \\
\midrule
\multicolumn{3}{c}{\textit{Training Settings}}\\
\midrule
\textbf{GPU} & $4\times8\times\text{H}20$ & $4\times8\times\text{H}20$ \\
\textbf{Optimizer} & AdamW & AdamW \\
\textbf{LR} & $1e-6$ & $1e-6$ \\
\textbf{Warmup Steps} & $10$ & $10$ \\
\textbf{Weight Decay} & $0.1$ & $0.1$ \\
\textbf{Entropy Coeff} & $0$ & $0$ \\
\textbf{Grad Clip} & $1.0$ & $1.0$ \\
\textbf{Loss Agg Mode} & token-mean & token-mean \\
\textbf{Clip Ratio Low} & $0.2$ & $0.2$ \\
\textbf{Clip Ratio High} & $0.28$ & $0.28$ \\
\textbf{KL in Reward} & False & False \\
\textbf{KL Coeff} & $0.0$ & $0.0$ \\
\textbf{KL Loss} & False & False \\
\textbf{KL Loss Coeff} & $0.0$ & $0.0$ \\
\textbf{Adv Estimator} & GRPO & GRPO \\
\textbf{N Responses / Prompt} & $16$ & $16$ \\
\textbf{Train BSZ} & $512$ & $512$ \\
\textbf{Mini BSZ} & $32$ & $32$ \\
\textbf{Gen BSZ} & $1024$ & $1024$ \\
\textbf{Filter Groups} & True & True \\
\textbf{Metric} & acc & acc \\
\textbf{Max Gen Batches} & $10$ & $10$ \\
\textbf{Overlong Buffer} & Enabled & Enabled \\
\textbf{Overlong Buffer Len} & $512$ & $512$ \\
\textbf{Penalty Factor} & $1.0$ & $1.0$ \\
\textbf{Top-p} & $1.0$ & $1.0$ \\
\textbf{Top-k} & $-1$ & $-1$ \\
\textbf{Temperature} & $1.0$ & $1.0$ \\
\textbf{Sampling} & Enabled & Enabled \\
\textbf{Offload} & False & False \\
\textbf{Use Dynamic BSZ} & True & True \\
\textbf{Max Prompt Len} & $1500$ & $512$ \\
\textbf{Max Resp Len} & $2596$ & $3584$ \\
\textbf{Initial Rollouts $N_1$} & $4$ & N/A \\
\textbf{Re-Prompting Rollouts $N_2$} & $3$ & N/A \\
\textbf{Top-$K$ Entropy} & $20$ & N/A \\
\midrule
\multicolumn{3}{c}{\textit{Testing Settings}}\\
\midrule
\textbf{Top-p} & $0.7$ & $0.7$ \\
\textbf{Top-k} & $-1$ & $-1$ \\
\textbf{Temperature} & $0.6$ & $0.6$ \\
\textbf{Max Prompt Len} & $512$ & $512$ \\
\textbf{Max Resp Len} & $3584$ & $3584$ \\
\bottomrule
\end{tabular}
\caption{Hyperparameter settings for CURE (First \& Second Stage) on Qwen-2.5-7B-Math}
\label{tab:dapo_qwen_hyperparams_twocol}
\end{table}
  \item \textbf{NFT}~\cite{chen2025bridging}, is a supervised, verifier-driven training scheme that builds an implicit negative policy from an LLM’s self-generated
    wrong answers while sharing parameters with the positive policy. By optimizing directly over all model generations using binary verifier feedback—without external teachers—NFT recycles failures into learning signals and removes the dependence on rejection sampling style supervision. Experiments on 7B and 32B math reasoning models show consistent gains over SL (Supervised Learning) baselines and parity or improvements versus leading RL methods
    such as GRPO and DAPO. Theoretically, NFT is equivalent to GRPO under strict on-policy training, helping
    bridge SL and RL in binary-feedback settings and opening a scalable path for self-improving LLMs.
  \item \textbf{Clip-CoV}~\cite{cui2025entropy} is an entropy-aware reinforcement learning update that clips token-wise gradients when the covariance between action probability
and logit change is high (an advantage-proportional
signal). By limiting confidence amplification on dominant tokens, it preserves policy entropy, sustains exploration, and prevents early entropy collapse. Clip-Cov is
plug-and-play with Policy Gradient variants, optimizer-agnostic, and incurs negligible overhead, prevents entropy collapse in multi-step reasoning.
\item \textbf{KL-CoV}~\cite{cui2025entropy} is a selective KL regularization that applies a KL penalty to tokens with high covariance, while leaving rare-but-promising actions less
constrained. This targeted control curbs overconfident
updates, maintains a healthy entropy trajectory, and
encourages exploration. KL-CoV complements global
KL control, works with common advantage functions,
and  prevents entropy collapse in reasoning
tasks.
\item \textbf{Beyond the 80/20 Rule}~\cite{wang2025beyond} presents an in-depth analysis of token‐level entropy patterns in chain‐of‐thought reasoning, showing that a small subset of high‐entropy “forking” tokens drives the model’s multi‐path exploration. By confining RLVR’s policy‐gradient updates to this critical 20 \% of tokens, Qwen28 achieves performance on the Qwen3-8B base model that is indistinguishable from full‐gradient training. \textbf{However, because the authors have not released the training code or any usable checkpoints, we did not include this work as a baseline.}
\end{itemize}

\subsection{Experiment Details}
Tab.~\ref{tab:dapo_qwen_hyperparams_twocol} presents the hyperparameters applied in both stages. In our baseline model, we adopt the same DAPO hyperparameters as those used in the two-stage setup.

\subsubsection{Detailed Explanation of CURE Hyperparameters}
In Tab.~\ref{tab:stage1-hparams} 4, we provide explanations for the special hyperparameters in CURE.
\begin{table*}[t]
\centering
\small
\begin{tabular}{l p{4.2cm} p{6.6cm}}
\hline
\textbf{Name}  & \textbf{Where in formulas} & \textbf{Role / Budget Impact} \\
\hline
Initial Rollouts $N_1$
& $\mathcal{G}(\mathbf{q},N_1)=\{\mathbf{o}_i\}_{i=1}^{N_1}\!\sim\!\pi_{\theta_{\text{old}}}(\cdot\mid \mathbf{q})$ (Eq.~\ref{eq:n1})
& Number of initial trajectories per query  $ \mathbf{q}_i$, adds $N_1$ samples.\\

Top-$K$ Entropy $K$
& $\mathcal{T}^{(i)}_K=\operatorname{TopK}_t(H_{i,t},K)$;\; $t_i^{\star}\!\sim\!\mathrm{Uniform}(\mathcal{T}^{(i)}_K)$ (Eq.~\ref{eq:t_star_full})
& Size of high-uncertainty candidate set for stochastic selection. \\

Re-Prompting Rollouts $N_2$
& $\mathcal{G}(\mathbf{q}'_i,N_2)=\{\mathbf{o}_{j}\}_{j=1}^{N_2}\!\sim\!\pi_{\theta_{\text{old}}}(\cdot\mid \mathbf{q}'_i)$ (Eq.~\ref{eq:new_rollouts_full})
& Number of trajectories per refined prompt $\mathbf{q}'_i$, adds $N_1*N_2$ re-prompted samples. \\
\hline
\end{tabular}
\caption{CURE hyperparameters and their occurrences in the corresponding formulas (Sec.~\ref{sec:cure-grpo}).}
\label{tab:stage1-hparams}
\end{table*}

\subsection{Evaluation Protocol}
Eurus-2-7B-PRIME-Zero, SimpleRL-Zero, Open-Reasoner-Zero, Oat-Zero, and LUFFY were retrieved from their official repositories and evaluated using each author’s recommended sampling parameters. NFT’s results are reported directly from its original paper, as neither the model nor its training code were released. In contrast, Clip-Cov and KL-CoV were trained in-house using the official training scripts provided by their authors with the initial model replaced by Qwen2.5-math-7b, since no pretrained checkpoints were available. To ensure the highest evaluation fidelity—and following the protocols established for Eurus-2-7B-PRIME-Zero and NFT we employed a combined verification pipeline (math-verify plus math-grader) and then manually reviewed every correct and incorrect sample across multiple test sets for a truly precise assessment.

\subsection{Word Cloud}
\begin{figure}[h]
  \centering
  \includegraphics[width=.98\columnwidth]{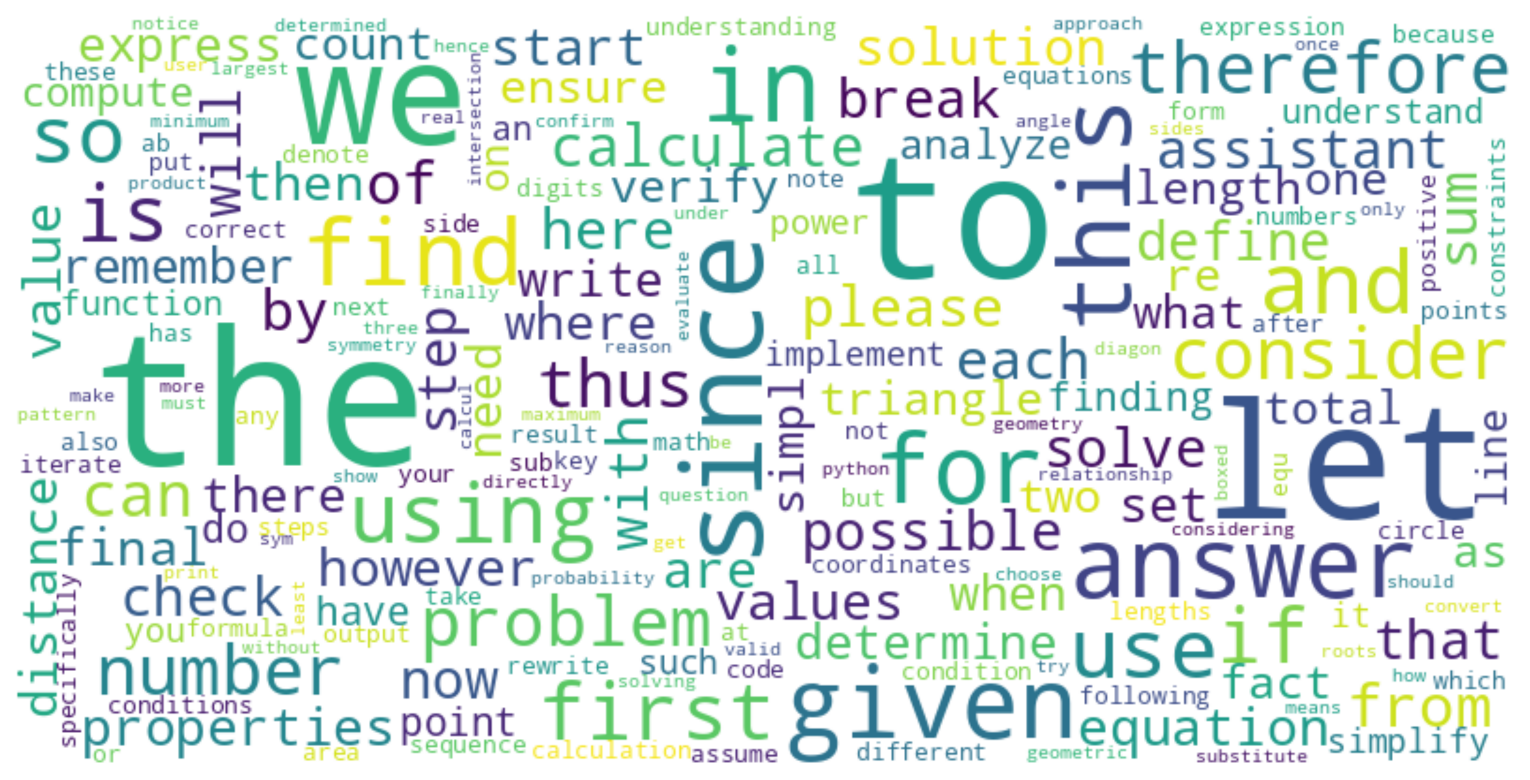}
  \caption{Word cloud generated by Qwen-2.5-Math-7B.}
  \label{fig:entropy_wordcloud_qwen}
  \vspace{-10pt}
\end{figure}

\begin{figure}[h]
  \centering
  \includegraphics[width=.98\columnwidth]{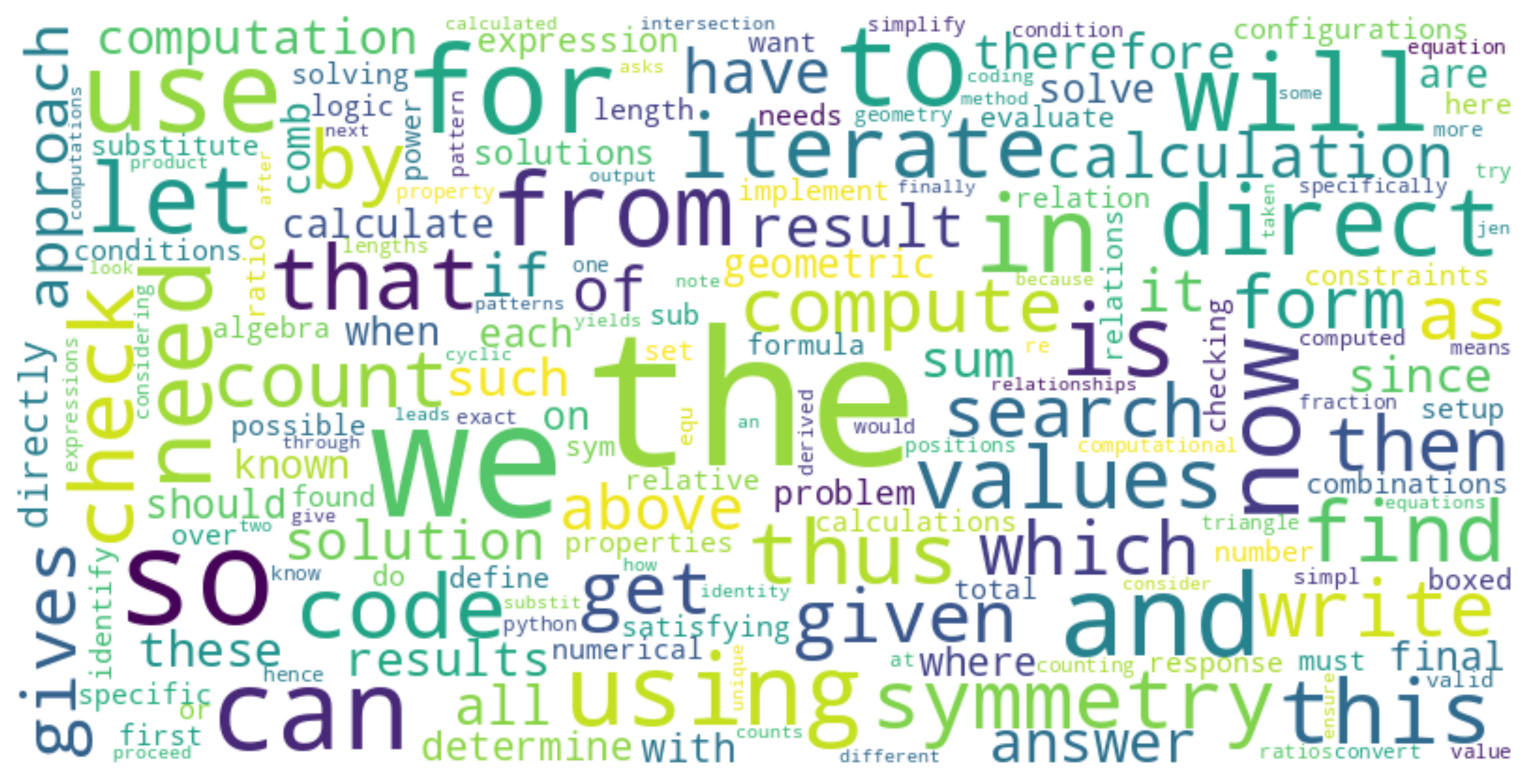}
  \caption{Word cloud generated by DAPO.}
  \label{fig:entropy_wordcloud_dapo}
  \vspace{-10pt}
\end{figure}

\begin{figure}[h]
  \centering
  \includegraphics[width=.98\columnwidth]{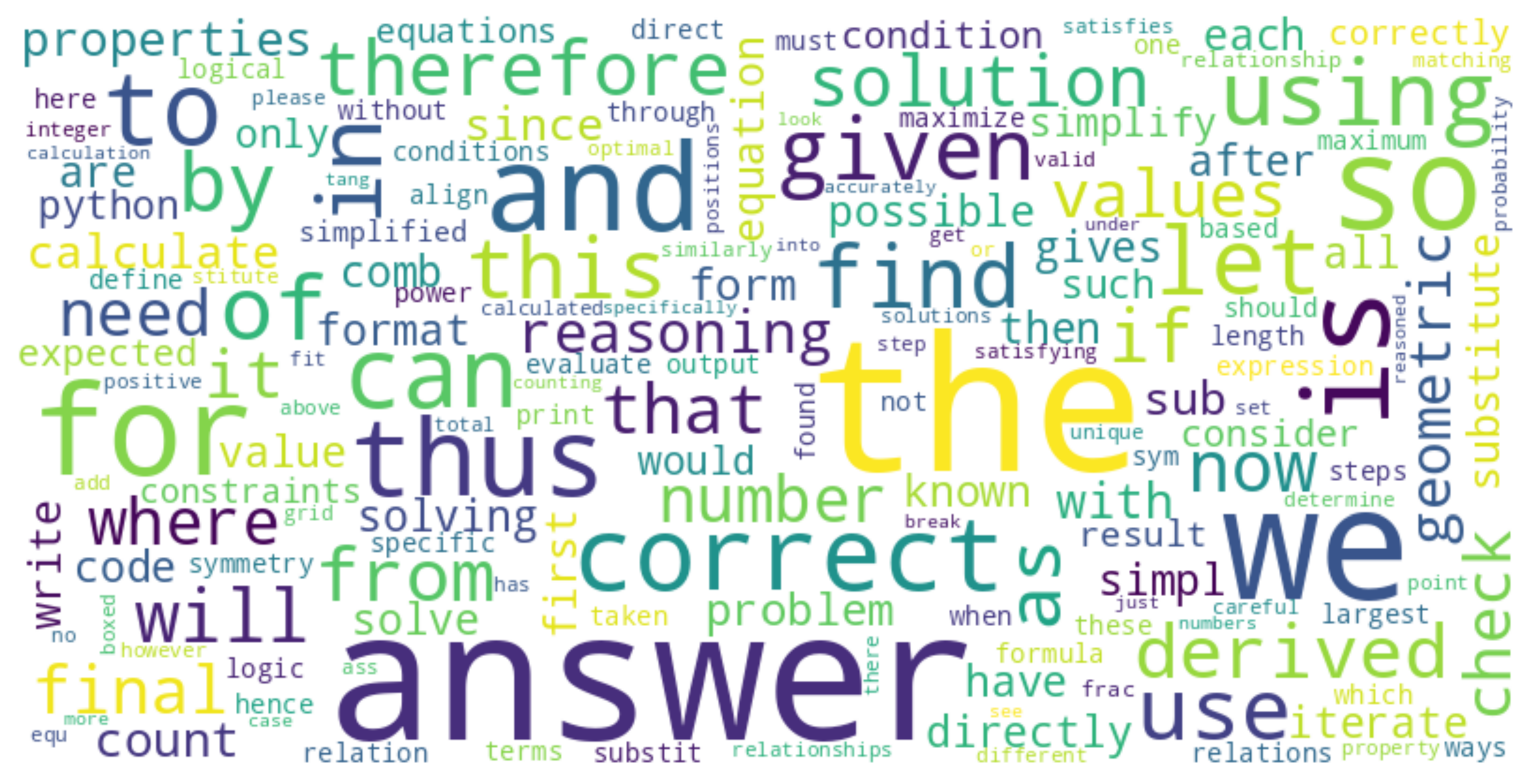}
  \caption{Word cloud generated by CURE First Stage.}
  \label{fig:entropy_wordcloud_cure}
  \vspace{-10pt}
\end{figure}

\subsection{Open Source}
Our code builds upon the foundations of VERL. To enhance the reproducibility of our work and support community development, we will release the model, training code, and evaluation code in the near future.

\section{Case Study}
\subsection{Template}
\[
\begin{aligned}
\textbf{System:}   &\; \text{Please reason step by step and enclose your }\\&\text{ final answer in } \backslash \text{boxed}\{\} \,.\\[6pt]
\textbf{User:}     &\; \{\,\text{Problem}\,\} \\[6pt]
\textbf{Assistant:}&\; \{\,\text{Answer}\,\}
\end{aligned}
\]
We used the simplest template with no special modifications.
\subsection{Policy Entropy Visualization}

\begin{figure}[h]
  \centering
  \includegraphics[width=.98\columnwidth]{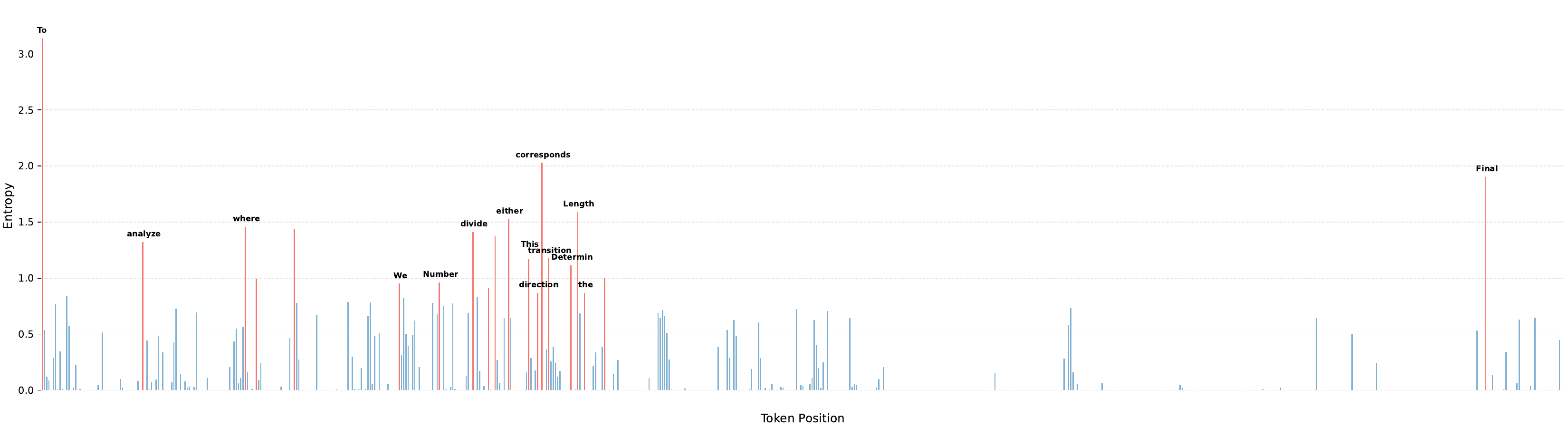}
  \caption{AIME24 Token-Level Entropy vs. Position (Qwen2.5-Math-7B) Case 1}
      \label{fig:entropy_pos_qwen25_1}
  \vspace{-10pt}

\end{figure}

\begin{figure}[h]
  \centering
  \includegraphics[width=.98\columnwidth]{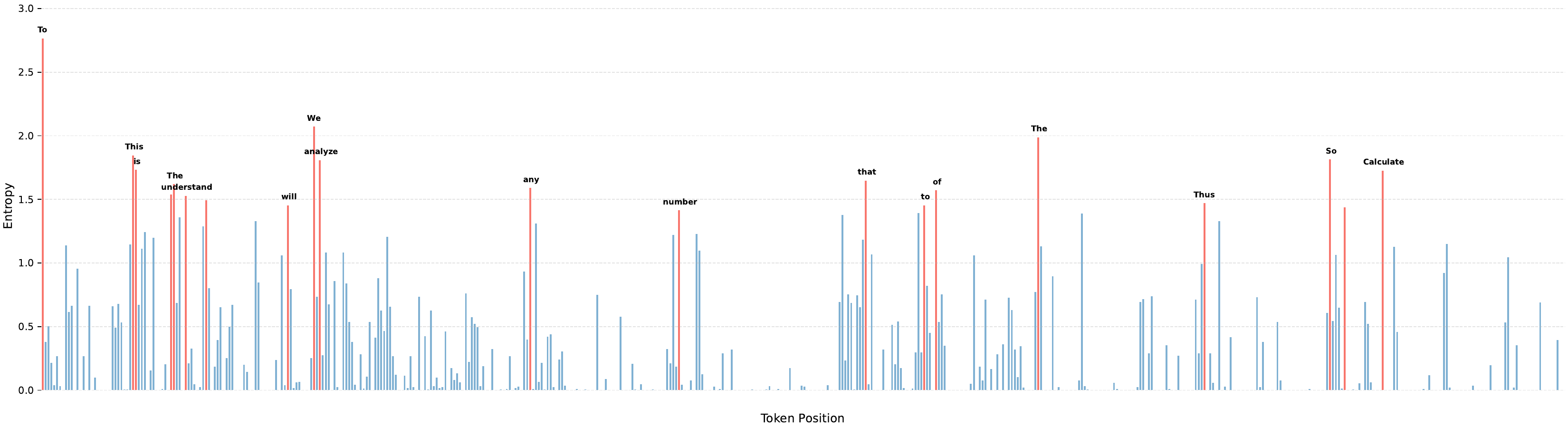}
  \caption{AIME24 Token-Level Entropy vs. Position (Qwen2.5-Math-7B) Case 2}
\label{fig:entropy_pos_qwen25_2}
  \vspace{-10pt}

\end{figure}
Since our initial model is Qwen2.5-Math-7B, we present here the variation of token-level entropy with token position in AIME24 in Fig.~\ref{fig:entropy_pos_qwen25_1} and Fig.~\ref{fig:entropy_pos_qwen25_2}. It can be observed that a large number of high-entropy tokens are either connectives or words that determine the direction of reasoning. These tokens exhibit relatively high entropy, which we attribute to the fact that their corresponding prefixes are more unfamiliar to the LLM. Therefore, we aim to treat the prefixes as new queries to enhance the model's exploratory capability.

\section{Limitation}
Limitations and Future Work
Due to computational constraints, our evaluation was confined to the Qwen family (up to the 7B variant), which restricts our understanding of cross-modal performance on larger or more diverse architectures. In future work, we will:
\begin{itemize}
  \item Scale to Qwen-2.5-VL: Extend our method to the upcoming Qwen-2.5-VL variant.
  \item Expand the coverage of the model: Apply the approach to additional families (e.g., LLaMA) and across varied size classes.
\end{itemize}
 \end{document}